\pdfoutput=1

\documentclass[11pt]{article}

\usepackage[table,xcdraw]{xcolor}
\usepackage[final]{acl}

\usepackage{times}
\usepackage{latexsym}
\usepackage{graphicx}
\usepackage{array}
\usepackage{booktabs}
\usepackage{graphicx, amsmath}
\usepackage{adjustbox}
\usepackage{caption}
\usepackage{multirow}
\usepackage{subcaption}
\usepackage{times}
\usepackage{latexsym}
\usepackage{natbib} 

\usepackage{xspace}
\newcommand{\proj}{LADs\xspace}

\usepackage{booktabs} 

\usepackage[most]{tcolorbox}

\usepackage{pifont}
\usepackage[T1]{fontenc}

\usepackage[utf8]{inputenc}

\usepackage{microtype}

\usepackage{inconsolata}

\usepackage{graphicx}

\title{LADs: Leveraging LLMs for AI-Driven DevOps}

\author{
    \textbf{Ahmad Faraz Khan\textsuperscript{1*}, Azal Ahmad Khan\textsuperscript{2*}, Anas Mohamed\textsuperscript{2}, Haider Ali\textsuperscript{1}, Suchithra Moonlinti\textsuperscript{2},} \\
    \textbf{Sabaat Haroon\textsuperscript{1}, Usman Tahir\textsuperscript{3}, Mattia Fazzini\textsuperscript{2}, Ali R. Butt\textsuperscript{1}, Ali Anwar\textsuperscript{2}} \\
    \textsuperscript{1}Virginia Tech, \textsuperscript{2}University of Minnesota, \textsuperscript{3} Ghulam Ishaq Khan Institute\\
    \texttt{\{ahmadfk, haiderali, sabaat, butta\}@vt.edu}\\
    \texttt{\{khan1069, moha1325, m0006012, mfazzini, aanwar\}@umn.edu}, \texttt{utahir243@gmail.com}\\
}

\begin{document}
\maketitle

\begin{abstract}

Automating cloud configuration and deployment remains a critical challenge due to evolving infrastructures, heterogeneous hardware, and fluctuating workloads. Existing solutions lack adaptability and require extensive manual tuning, leading to inefficiencies and misconfigurations. We introduce \proj, the first LLM-driven framework designed to tackle these challenges by ensuring robustness, adaptability, and efficiency in automated cloud management. Instead of merely applying existing techniques, \proj provides a principled approach to configuration optimization through in-depth analysis of what optimization works under which conditions. By leveraging Retrieval-Augmented Generation, Few-Shot Learning, Chain-of-Thought, and Feedback-Based Prompt Chaining, \proj generates accurate configurations and learns from deployment failures to iteratively refine system settings. Our findings reveal key insights into the trade-offs between performance, cost, and scalability, helping practitioners determine the right strategies for different deployment scenarios. For instance, we demonstrate how prompt chaining-based adaptive feedback loops enhance fault tolerance in multi-tenant environments and how structured log analysis with example shots improves configuration accuracy. Through extensive evaluations, \proj reduces manual effort, optimizes resource utilization, and improves system reliability. By open-sourcing \proj, we aim to drive further innovation in AI-powered DevOps automation.


\end{abstract}

\renewcommand{\thefootnote}{}
\footnotetext{* Equal contributions (ordered via coin-flip).}

\section{Introduction}
\label{sec:Intro}
Cloud-based systems involve complex deployment and continuous maintenance, which require comprehensive expertise in software development, systems engineering, security, networking, data, operations management, and domain-specific knowledge. This complexity and the continuous need for configuration updates can be time-consuming and expensive for startups that rely on computing, storage, data processing, and ML systems. 


At large-scale companies like Meta, thousands of configuration file “diffs” are committed daily. These changes outpace code updates and misconfigurations remain a top cause of production incidents~\cite{lian2024configuration}. The rapid pace of configuration changes in modern cloud applications shows a pressing need for automated intelligent solutions that can handle dynamic environments with minimal human intervention.

A recent study indicates that $77\%$ of organizations rely on DevOps to deploy their services~\cite{cloudzero2024devops}. Yet many struggle with effective management because DevOps teams spend significant time on observability, monitoring, and security tasks, which consume $54\%$ of their efforts. Only three out of ten organizations fully understand their cloud spendings~\cite{cloudzero2024devops}, highlighting resource-budgeting challenges that grow more complex in heterogeneous architectures hosting distributed services for data processing, caching, and ML workloads.

Configuration and deployment form the core of cloud service management. Still, existing solutions often remain system-specific and fail to generalize across different services, leading to extensive manual tuning and frequent misconfigurations. For example, tuning an LSM-based key-value store like RocksDB involves over 100 parameters~\cite{lao2024gptuner,thakkar2024can}, while large-scale platforms like Spark and Dask require careful optimization of parameters such as \texttt{spark.executor.memory} and \texttt{spark.core.max} to meet performance goals~\cite{zheng2021multi}. Engineers often build validators for these large parameter spaces, yet many remain uncovered due to cost and complexity~\cite{lian2024configuration}. Such issues are compounded by dynamic workloads, evolving user goals, and the need for real-time reconfiguration to maintain reliability and efficiency.
Automated monitoring and maintenance are also essential to avoid failures caused by version mismatches, poor memory allocation or misconfigured network policies~\cite{zhang2021understanding}. DevOps engineers often rely on tools like Prometheus~\cite{Prometheus} for observability, yet these tasks demand significant time and are complicated by limited visibility into operational costs.

Recent advances in Large Language Models (LLMs) offer a way to streamline these processes. Cloud providers such as AWS~\cite{aws_cloud}, IBM Cloud~\cite{ibm_cloud}, and GCP~\cite{gcp} could leverage agentic LLM-based solutions to automate their configuration and deployment pipelines, thereby reducing human intervention. However, developing these solutions introduces challenges like integrating domain-specific knowledge, accurate configuration generation, robust debugging, and adaptation to heterogeneous environments.

In this work, we propose an agentic LLM framework called LADS that overcomes the challenges in automating the configuration and deployment of distributed cloud services. Our contributions are as follows: 1) A novel LLM-based approach that unifies Instruction Prompting (IP), hyperparameter tuning, Few-Shot Learning (FSL), Retrieval-Augmented Generation (RAG), and Chain of Thought (CoT) based on the analyzed strength of each technique for automated DevOps. 2) A feedback-based prompt-chaining method for maintenance that sends error logs and stage details back to the LLM agent to improve reliability. 3) An approach that selectively incorporates domain-specific knowledge (via retrieval-augmented generation) and user intent (via instruction prompting and chain of thought) to reduce cost and mitigate context-window limitations. 4) We also provide the first-of-its-kind benchmark and dataset for static and dynamic validation of automated DevOps and assessing alignment with user intent. 5) An analysis of trade-offs between performance gains and operational costs, offering guidance to practitioners seeking to optimize LLM-based deployments at scale.
{\proj} is designed to reduce the burden on DevOps teams by providing an intelligent and scalable framework that lowers misconfiguration risks and ensures robust cloud application management.
\section{Related Work}
\label{sec:RelatedWork}

LLMs have been explored for various configuration tasks, but existing approaches often address narrow, isolated challenges rather than providing an adaptive, end-to-end solution. Kannan et al.\cite{kannan2023software} employ LLMs for software tool configuration, particularly hyperparameter tuning in ML applications. Similarly, PreConfig\cite{li2024preconfig} applies text-to-text transformations for network configuration but lacks adaptability to dynamic cloud environments. Mondal et al.~\cite{mondal2023router} demonstrate the potential of LLMs in router configuration but highlight their susceptibility to critical errors, necessitating manual validation. Unlike these methods, {\proj} is designed for cloud system deployment, integrating continuous feedback and adaptive optimization to refine configurations in real time, reducing inefficiencies and manual intervention.

LLMs have also been explored for debugging and validation. Fu et al.\cite{fu2024missconf} and Lian et al.\cite{lian2024ciri} use LLMs to identify misconfigurations, but their approaches are reactive, focusing on error detection post-deployment rather than preventing misconfigurations from occurring. Similarly, IaC-Eval~\cite{kon2024iac-eval} introduces a benchmark for LLM-generated Infrastructure-as-Code (IaC), revealing challenges in static correctness evaluation but lacking mechanisms for adaptive tuning. {\proj} differs by taking a proactive approach—leveraging Retrieval-Augmented Generation (RAG), Feedback-Based Prompt Chaining, and structured logs to iteratively refine configurations before deployment, reducing the likelihood of failures rather than just identifying them.

Beyond configuration validation, Ruiz~\cite{ruiz2024software} and Malul~\cite{malul2024genkubesec} propose post-deployment strategies for automated software improvement and misconfiguration detection, respectively. While these methods enhance system robustness, they primarily respond to issues rather than optimizing configurations dynamically. In contrast, {\proj} integrates Few-Shot Learning, Chain-of-Thought reasoning, and structured feedback loops to continuously refine and optimize cloud configurations throughout the deployment lifecycle, balancing performance, cost, and reliability.

Unlike prior work that focuses on specific aspects of configuration, validation, or debugging, {\proj} is the first holistic, LLM-driven framework to strategically combine adaptive optimization, structured feedback, and intelligent automation for cloud system deployment. By addressing configuration correctness, resource efficiency, and deployment robustness, {\proj} sets a new benchmark for intelligent DevOps automation.

\definecolor{pink}{HTML}{EAD5DC}
\begin{table}[h]
    \centering
    \footnotesize
    \begin{tabular}{lccccc}
        \noalign{\hrule height 1pt}
        \rowcolor{pink}
        \textbf{Metric} & \textbf{Redis} & \textbf{Ray} & \textbf{Dask}\\
        \hline
        \textbf{Max Keys}      & \textbf{221}  & 110  & 192   \\
        \textbf{Max Depth}     & 7    & \textbf{8}    & 7    \\
        \textbf{Avg Keys}      & 77 ± 44   & 44 ± 20   & \textbf{82 ± 35}  \\
        \textbf{Avg Depth}     & 5.0 ± 0.52  & \textbf{5.5 ± 0.68}  & 5.1 ± 0.50  \\
        \textbf{No. of Features} & \textbf{19} & 15 & 16 \\
        \noalign{\hrule height 1pt}
    \end{tabular}
    \caption{Comparison of Configuration Complexity based on number of configurable parameter keys, depth of nested parameters, and features tested.}
    \label{tab:config_complexity}
\end{table}

\section{Benchmark and Dataset}
\label{sec: Benchmark and Dataset}

Our validation approach is twofold. We adopt static and dynamic validation techniques from Software Engineering~\cite{static_dynamic_validation}. The static technique checks whether the generated configuration aligns with the user intent, similar to IaC-Eval~\cite{kon2024iac-eval}. The dynamic technique deploys the generated configuration and measures its performance, cost, or scale using benchmark applications to confirm that the defined user intent is satisfied.

\emph{For the static validation} of {\proj}, we required a suitable benchmark. However, to our knowledge, no public datasets exist that validate user intent, deployment success, and performance for cloud application configuration. \emph{To address this gap, we created a human-curated dataset that we plan to release publicly.} This dataset tests configuration generation for distributed systems like Dask, Redis, and Ray, and we continue to add more systems. We chose these systems because of their popularity (based on GitHub stars) and because they represent different microservice tasks including computing, caching, storage, and ML. They also offer varying configuration complexity levels, as shown in Table~\ref{tab:config_complexity}. The Table compares the configuration complexity of Redis, Ray, and Dask based on the max and average number of configurable key parameters, the max and average depth of nested configurable parameters, and the number of features that were evaluated for each distributed system in our dataset. Redis has the highest Max Keys (221) and Avg Keys (77 ± 44), indicating its vast configuration space and it is also the most rigorously tested with 19 features tested in our dataset, while Ray has the deepest structure (Max Depth = 8, Avg Depth = 5.5 ± 0.68), suggesting more hierarchical complexity. Dask offers a balanced profile (192 keys, 16 features), making it relatively easy to configure.

The dataset itself includes user intents, prompts, and validators for each distributed service, with 60 data points per service. These 60 configurations per service are categorized into six difficulty levels, ranging from beginner-friendly tasks (Level 1) to expert-level challenges (Level 6). The prompts in the dataset define the high-level task, such as: ``\textit{Create a basic Dask cluster for deployment on Kubernetes in the namespace ‘dask,’ with scheduler name ‘dask-scheduler.’}'' Where as, the user intents specify precise requirements, e.g., ``\textit{The Dask scheduler should include one scheduler and two worker nodes.}'' During evaluation with this benchmark, the LLM agent takes the user intent and prompt as input and generates an output configuration. The Python-based validator functions provide a fine-grained method to verify whether the generated configuration aligns with the prompt and the user intent. Unlike prior works such as IaC-Eval~\cite{kon2024iac-eval}, which use regex-based validation, we choose Python for its flexibility and better alignment with LLM capabilities~\cite{zhong2024ldb, hu2025qualityflow, huang2023agentcoder}. Since LLMs excel at generating Python-based verifiers~\cite{zhong2024ldb, liang2024improving}, this approach also simplifies future automation of validator generation using LLM agents.\looseness=-1

For the dynamic validation, we deploy the system in multi-tenant cloud environments and assess its performance through official benchmark applications from each system’s documentation~\cite{Dask_Scaling_2017,Redis_Benchmarks,KubeRay_GitHub}. We use a containerized setup with Kubernetes, commonly employed for distributed microservices like Dask, Redis, and Ray. We define three user intents widely used in related DevOps studies~\cite{DevOps_intents}: “Low cost and resource consumption,” “High scalability,” and “High efficiency balancing processing costs, resource consumption, and scalability.” The first two focus on simpler metrics such as latency, cost, and resource usage, while the third presents a complex trade-off scenario. These intents are provided to the LLM agent, which generates a configuration and deploys the system on Kubernetes. We measure alignment with the user intent by running the official benchmarks while tracking completion time, cost, and resource consumption. The deployment is continuously monitored to validate if the configuration meets the user's performance criteria.

\section{Method}
\label{sec: Design}

Our solution offers an LLM-driven method to simplify the configuration management of distributed systems on the cloud. By leveraging multiple LLMs and Infrastructure as Code (IaC), users can define requirements in natural language and reduce complexity without deep technical expertise. Cloud applications vary widely in their configuration parameters, making it infeasible to train specialized models for each application. Instead, we \emph{treat LLMs as black-box modules and create a generic solution capable of configuring a broad range of cloud applications.} However, utilizing generative models as is for configuration generation is insufficient due to the inherent gap between the structure of configuration requirements and the nature of LLM inference. To bridge this gap, we focus on enhancing configuration generation through inference-phase optimizations. Implementing these optimizations presents additional challenges, which we address in the next section, providing the foundation for the design of our final solution.\looseness=-1

\begin{figure}
  \begin{center}
    \includegraphics[width=\columnwidth]{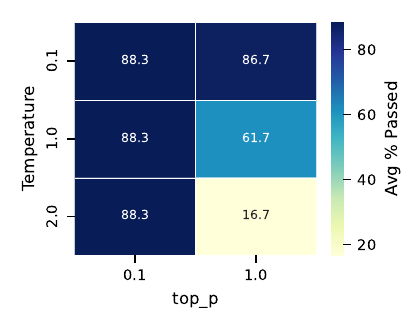}
    \vspace{-2.5em}
    \caption{Average Performance across all difficulty 
    levels with varying Temperature and top\_p parameters.}
    \label{fig:temperature_topp}
  \end{center}
    \vspace{-1.5em}
  
\end{figure}

\subsection{Structural integrity of configurations}
\label{subsec: Structural integrity of configurations}

The main challenge arises from the non-deterministic nature of generative models~\cite{non_determenism}. A single prompt can generate a different response each time, which complicates consistent extraction of the information needed to form a syntactically correct configuration file. This requires controlling model randomness and cleaning the generated responses.

\paragraph{Pre-processing Inference.}
Since LLM can generate responses in various formats, we preprocess its output by parsing and removing any extraneous content that could break the structure of a YAML file. This step strips out explanations, reasoning, or examples commonly included in the model's response. After this initial cleaning, we move on to controlling the model's randomness.

\paragraph{Controlling model randomness.}
The inference pipeline includes several parameters that affect the quality and diversity of generated outputs. Among these, \emph{temperature} and \emph{top\_p} (nucleus sampling) are especially important. The \emph{temperature} parameter scales the logits before softmax: a lower value yields more deterministic outputs by biasing toward high-probability tokens, while a higher value fosters diversity. The \emph{top\_p} parameter selects the smallest set of tokens whose cumulative probability exceeds \emph{p}, focusing on the most likely tokens yet retaining some variety.
Figure~\ref{fig:temperature_topp} shows how performance degrades (becomes non-deterministic) with higher \emph{temperature} and \emph{top\_p} values. To achieve more accurate and deterministic results in configuration generation, we keep these parameters lower than $0.1$.\looseness=-1





\paragraph{Instruction Prompting (IP).} Apart from parameter optimization, instruction prompting~\cite{wei2021finetuned} is vital for maintaining the structural integrity of generated YAML files. Figure~\ref{fig:prompt_eng_example} in the Appendix~\ref{subsec: Examples of optimizations in LADs} illustrates an engineered prompt that instructs the LLM to produce only relevant configuration details. Without IP, we often see severe inconsistencies that render the YAML files unusable, even with pre-processing. When we rely solely on low \emph{top\_k} with minimal data pre-processing, the outputs fail structural validation before we can assess conformance. After applying IP, data pre-processing, and randomness control, structural errors become negligible. As shown in Table~\ref{tab:main_table}, IP helps the models pass some prompt-alignment checks, however, performance remains insufficient for fully automated DevOps pipelines.




\subsection{Context Provision}
\label{Context Provision}

While techniques like output pre-processing and prompt optimization with randomness control can yield syntactically correct configuration files, they do not address deployment complexity under varying resource conditions, workload dynamics, or changing user intents. They also lack monitoring and maintenance capabilities. Configuration generation requires contextual data that includes system information, resource limits, application documentation, and user intent. We categorize this context as either \textit{structured} (e.g., reference outputs and configuration examples) or \textit{unstructured} (e.g., application documentation, user intent, and system data). Common methods for incorporating context include RAG~\cite{lewis2020retrieval}, few-shot~\cite{brown2020language_few_shot,gao2021making_few_shot}, and chain of thought~\cite{wei2022chainofthought}. We test all of these alongside instruction prompting, pre-processing, and randomness control to see how context information, passed through these techniques, affects configuration generation.


\paragraph{Retrieval Augmented Generation (RAG).}
We first evaluate the RAG optimization by providing structured and unstructured information via a stuff chain, FAISS retriever~\cite{johnson2019billionFAISS}, a 5000-token chunk size, and a 500-token overlap. We also include the instruction prompting techniques described in~\S\ref{subsec: Structural integrity of configurations} to ensure structurally valid YAML files. Table~\ref{tab:main_table} compares this approach to the instruction prompting optimization. The performance gains are minimal because structured schema information loses coherence when broken into chunks, and there is a risk of introducing fragmented or noisy content~\cite{noisyRAG}. Moreover, RAG does not guarantee the retrieval of essential chunks, so if the retrieved chunk contains the configuration schema, critical configuration formatting details may be lost. This explains why we see only modest improvements over IP alone. These findings lead us to a key observation: \textbf{OB\ding{182} RAG is better suited for passing unstructured context information that can be filtered and utilized in the form of relevant chunks of information required. This reduces the amount of data passed via input saving valuable input tokens for other structured information.}
To resolve this issue, we next consider passing structured configuration schema context via a more direct approach through the few-shot technique.


\paragraph{Few-Shot Learning (FSL).}
For the few-shot technique, we draw on prior work in structured code generation~\cite{lian2024ciri,kon2024iac-eval}, which demonstrates that providing example shots significantly improves performance. The LLM agent benefits from having a clear reference for both the schema and the type of content expected. We supply in-context few-shot examples of prompts, user intents, and both correct and incorrect outputs, as illustrated in Figure~\ref{fig:few_shot_example} in the Appendix~\ref{subsec: Examples of optimizations in LADs}. Table~\ref{tab:main_table} confirms that few-shot examples lead to substantial gains in accuracy. This success is driven by the LLM’s strong in-context learning ability, which helps it better interpret configuration requirements and user intents.


\paragraph{Chain-of-Thought (CoT).}


\begin{figure}
  \begin{center}
    \includegraphics[width=\columnwidth]{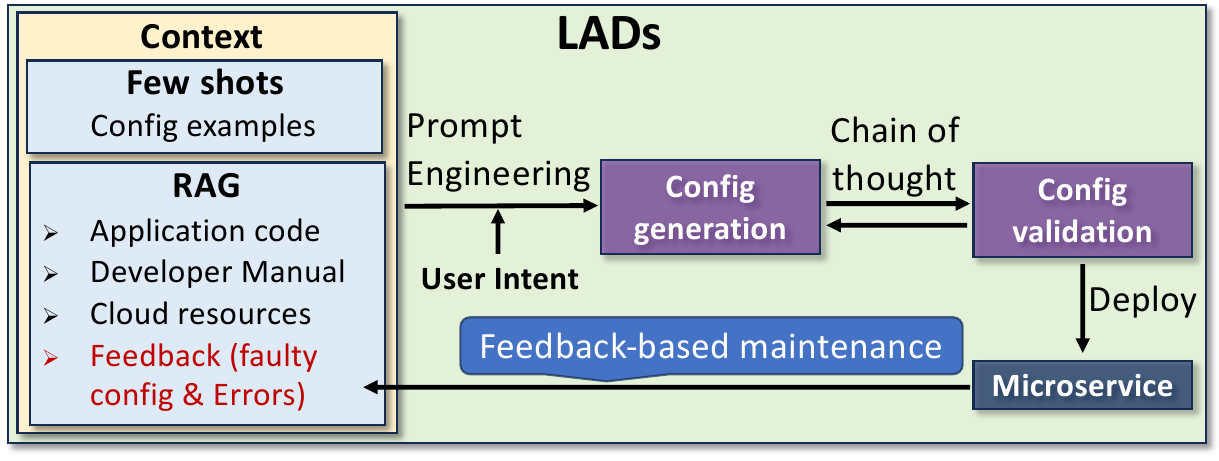}
    \caption{High-level overview of \proj for automated cloud application management. The system leverages LLM agents and integrates in-context few-shot and instruction prompting with RAG and prompt chaining.}
    \label{fig:overview}
  \end{center}
  \vspace{-20pt}
\end{figure}


For chain-of-thought optimization, we adopt the zero-shot learning method from related work~\cite{wei2022chainofthought}. This method prompts the LLM to reason in steps by inserting instructions such as “Let’s think step by step,” after each response. Table~\ref{tab:main_table} shows only modest improvements with this approach. The limited gains arise because the LLM lacks sufficient domain-specific knowledge (e.g., system specifications and configuration schema) to leverage multi-step reasoning for configuration generation.

By contrast, the most significant improvements come from providing example shots as input, giving the model a structural reference for valid configurations. These findings leads us to a key observation: \textbf{OB\ding{183} LLMs needs a more concrete schema or format to guide its outputs. Few-Shot and CoT methods are better suited for passing structured information. Few-Shot examples help models learn the configuration format and structure, while CoT enables them to correct issues like overlooked user intent.}

\paragraph{\proj.}
We combine multiple LLM-based optimization methods to incorporate the necessary context while balancing cost-effectiveness. However, our novelty lies in identifying the strengths of each optimization and using them appropriately in our system's design. Our design focuses on two overarching goals: maximizing the accuracy and efficiency of LLM-driven configuration and deployment and minimizing token-related expenses. Figure~\ref{fig:overview} shows the methodology applied in LADs. Apart from the structural integrity measures discussed in \S~\ref{subsec: Structural integrity of configurations}, to achieve accurate configuration generation aligned with the user intent, we rely on a combination of instruction prompting, chain-of-thought, RAG, hyperparameter tuning, and prompt chaining. Based on the insights gained from the individual application of each optimization, we use the \emph{few-shot} method to provide information that requires structure preservation such as the user intents (for example, “prioritize cost savings” or “maximize throughput under 4 CPU cores”) and \emph{structured configuration examples} which we call shots. These few-shot examples as shown in Figure~\ref{fig:few_shot_example} of Appendix~\ref{subsec: Examples of optimizations in LADs} show critical parameters, and configuration format across different systems to help the LLM agent understand how the response should look. The large chunks of information such as Kubernetes cluster specs or extensive application manuals are passed to the input prompt via \emph{Retrieval-Augmented Generation (RAG)} to fetch only relevant chunks at run time, thus avoiding prompt overload and reducing token costs.

A key component of \proj is ongoing monitoring and maintenance, where the LLM identifies and addresses configuration or deployment failures. As shown in Figure~\ref{fig:overview}, for statically verifying the configuration in \proj during configuration generation, we rely on \emph{Chain-of-thought} reasoning, which reasons if the model output is aligned with the user intent, prompt, and shots to ensure correctness. For dynamic validation, \proj includes a \emph{feedback-based prompt chaining loop} as shown in Figure~\ref{fig:overview}. This feedback-based prompt chaining supplies the model with filtered logs, error messages, warnings, and stage-specific failure points. For example, when a system passes syntax checks but fails during deployment (due to a reason such as ``deployment failed after static validation succeeded due to insufficient memory''), that detail is fed back into the prompt to help narrow down the root cause.\looseness=-1

\emph{By strategically utilizing the strengths of instruction prompting, hyperparameter tuning, few-shot learning, chain-of-thought, and RAG, \proj tackles configuration complexity, deployment complexity, dynamic user goals, and maintenance. This consolidated approach provides cost efficiency, adaptability, and performance alignment with user intent in an LLM-driven cloud deployment pipeline.} \proj also ensures that the token usage is under control without sacrificing performance allowing it to adapt quickly to dynamic workloads or user objectives while maintaining both affordability and efficiency.

\renewcommand{\arraystretch}{1.2}
\definecolor{pink}{HTML}{EAD5DC}
\definecolor{blue}{HTML}{EEE3E7}
\definecolor{lightgreen}{HTML}{E8F0E6}
\definecolor{lightred}{HTML}{F4E8E6}

\renewcommand{\arraystretch}{1.2}
\definecolor{pink}{HTML}{EAD5DC}
\definecolor{blue}{HTML}{EEE3E7}
\definecolor{lightgreen}{HTML}{E8F0E6}
\definecolor{lightred}{HTML}{F4E8E6}

\begin{table}[t]
\centering
\small
\setlength{\tabcolsep}{2pt} 
\renewcommand{\arraystretch}{1.0} 
\begin{tabular*}{\linewidth}{@{\extracolsep{\fill}}p{1.5cm}|p{1.1cm}p{1.1cm}p{1.1cm}p{1.1cm}p{1cm}}
\noalign{\hrule height 1pt} 
\rowcolor{pink}
\textbf{Models} & \textbf{IP} & \textbf{CoT} & \textbf{RAG} & \textbf{FSL} & \textbf{LADs} \\
\noalign{\hrule height 1pt} 
\rowcolor{blue} 
\multicolumn{6}{c}{\textbf{Dask}} \\
\noalign{\hrule height 1pt} 
\textbf{CL-7B}   & 5.00  & 0.00 \textcolor{red}{$\downarrow$}  & 6.67 \textcolor{green}{$\uparrow$}  & 28.34 \textcolor{green}{$\uparrow$}  & \textbf{33.34} \textcolor{green}{$\uparrow$} \\
\textbf{CL-13B}  & 5.00  & 3.34 \textcolor{red}{$\downarrow$}  & 8.34 \textcolor{green}{$\uparrow$}  & 43.34 \textcolor{green}{$\uparrow$}  & \textbf{48.34} \textcolor{green}{$\uparrow$} \\
\textbf{L3.1-8B}  & 18.34 & 8.34 \textcolor{red}{$\downarrow$}  & 18.34 \textcolor{green}{$\uparrow$} & \textbf{46.67} \textcolor{green}{$\uparrow$} & 40.00 \textcolor{green}{$\uparrow$} \\
\textbf{DS2-16B} & 8.34  & 20.00 \textcolor{green}{$\uparrow$} & 10.00 \textcolor{green}{$\uparrow$} & 48.34 \textcolor{green}{$\uparrow$} & \textbf{50.00} \textcolor{green}{$\uparrow$} \\
\textbf{Q2.5-7B}  & 21.67 & 16.67 \textcolor{red}{$\downarrow$} & 15.00 \textcolor{red}{$\downarrow$} & 55.00 \textcolor{green}{$\uparrow$} & \textbf{56.67} \textcolor{green}{$\uparrow$} \\
\textbf{Q2.5-14B} & 48.34 & 45.00 \textcolor{red}{$\downarrow$} & 18.34 \textcolor{red}{$\downarrow$} & 56.67 \textcolor{green}{$\uparrow$} & \textbf{\underline{68.34}} \textcolor{green}{$\uparrow$} \\
\noalign{\hrule height 1pt} 
\rowcolor{blue} 
\multicolumn{6}{c}{\textbf{Redis}} \\
\noalign{\hrule height 1pt} 
\textbf{CL-7B}   & 16.67 & 1.67 \textcolor{red}{$\downarrow$}  & 16.67 \textcolor{green}{$\uparrow$} & 26.67 \textcolor{green}{$\uparrow$} & \textbf{28.34} \textcolor{green}{$\uparrow$} \\
\textbf{CL-13B}  & 15.00 & 1.67 \textcolor{red}{$\downarrow$}  & 13.33 \textcolor{red}{$\downarrow$} & \textbf{25.00} \textcolor{green}{$\uparrow$} & 23.34 \textcolor{green}{$\uparrow$} \\
\textbf{L3.1-8B}  & 21.67 & 23.34 \textcolor{green}{$\uparrow$} & 21.67 \textcolor{green}{$\uparrow$} & 35.00 \textcolor{green}{$\uparrow$} & \textbf{36.67} \textcolor{green}{$\uparrow$} \\
\textbf{DS2-16B} & 20.00 & 25.00 \textcolor{green}{$\uparrow$} & 23.34 \textcolor{green}{$\uparrow$} & 30.00 \textcolor{green}{$\uparrow$} & \textbf{38.34} \textcolor{green}{$\uparrow$} \\
\textbf{Q2.5-7B}  & 20.00 & 20.00 \textcolor{green}{$\uparrow$} & 16.67 \textcolor{red}{$\downarrow$} & 30.00 \textcolor{green}{$\uparrow$} & \textbf{31.67} \textcolor{green}{$\uparrow$} \\
\textbf{Q2.5-14B} & 45.00 & 43.34 \textcolor{red}{$\downarrow$} & 31.67 \textcolor{red}{$\downarrow$} & 31.67 \textcolor{red}{$\downarrow$} & \textbf{\underline{58.34}} \textcolor{green}{$\uparrow$} \\
\noalign{\hrule height 1pt} 
\rowcolor{blue} 
\multicolumn{6}{c}{\textbf{Ray}} \\
\noalign{\hrule height 1pt} 
\textbf{CL-7B}   & 3.34  & 0.00 \textcolor{red}{$\downarrow$}  & 1.67 \textcolor{red}{$\downarrow$} & 35.00 \textcolor{green}{$\uparrow$}  & \textbf{41.67} \textcolor{green}{$\uparrow$} \\
\textbf{CL-13B}  & 0.00  & 1.67 \textcolor{green}{$\uparrow$}  & 0.00 \textcolor{green}{$\uparrow$} & 35.00 \textcolor{green}{$\uparrow$}  & \textbf{41.67} \textcolor{green}{$\uparrow$} \\
\textbf{L3.1-8B}  & 3.34  & 3.34 \textcolor{green}{$\uparrow$}  & 3.34 \textcolor{green}{$\uparrow$} & 51.67 \textcolor{green}{$\uparrow$}  & \textbf{56.67} \textcolor{green}{$\uparrow$} \\
\textbf{DS2-16B} & 5.00  & 5.00 \textcolor{green}{$\uparrow$}  & 13.34 \textcolor{green}{$\uparrow$} & 28.34 \textcolor{green}{$\uparrow$}  & \textbf{30.00} \textcolor{green}{$\uparrow$} \\
\textbf{Q2.5-7B}  & 20.00 & 20.00 \textcolor{green}{$\uparrow$} & 18.33 \textcolor{red}{$\downarrow$} & \textbf{61.67} \textcolor{green}{$\uparrow$} & 53.34 \textcolor{green}{$\uparrow$} \\
\textbf{Q2.5-14B} & 30.00 & 26.67 \textcolor{red}{$\downarrow$} & 18.34 \textcolor{red}{$\downarrow$} & 63.34 \textcolor{green}{$\uparrow$}  & \textbf{\underline{70.00}} \textcolor{green}{$\uparrow$} \\
\noalign{\hrule height 1pt}
\end{tabular*}
\caption{\textbf{Performance of Various LLMs on Multiple Applications with Different Optimizations.} The highest-performing optimization for each LLM is bolded, while the overall best-performing pair is bolded and underlined. Abbreviations: \textbf{CL-7B} (CodeLlama7b), \textbf{CL-13B} (CodeLlama13b), \textbf{L3.1-8B} (Llama 3.1 8b), \textbf{DS2-16B} (DeepSeekV2 16b), \textbf{Q2.5-7B} (Qwen2.5 7b), \textbf{Q2.5-14B} (Qwen2.5 14b). Arrows indicate change from IP: green ($\uparrow$) for improvements and red ($\downarrow$) for decreases.}
\vspace{-2em}
\label{tab:main_table}
\end{table}

\begin{figure*}[ht]
  \centering
      \includegraphics[width=0.8\linewidth]{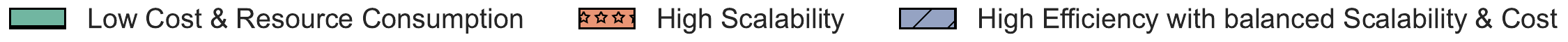}
    \hfill
  \begin{subfigure}[t]{0.49\textwidth}
    \centering
    \includegraphics[width=\linewidth]{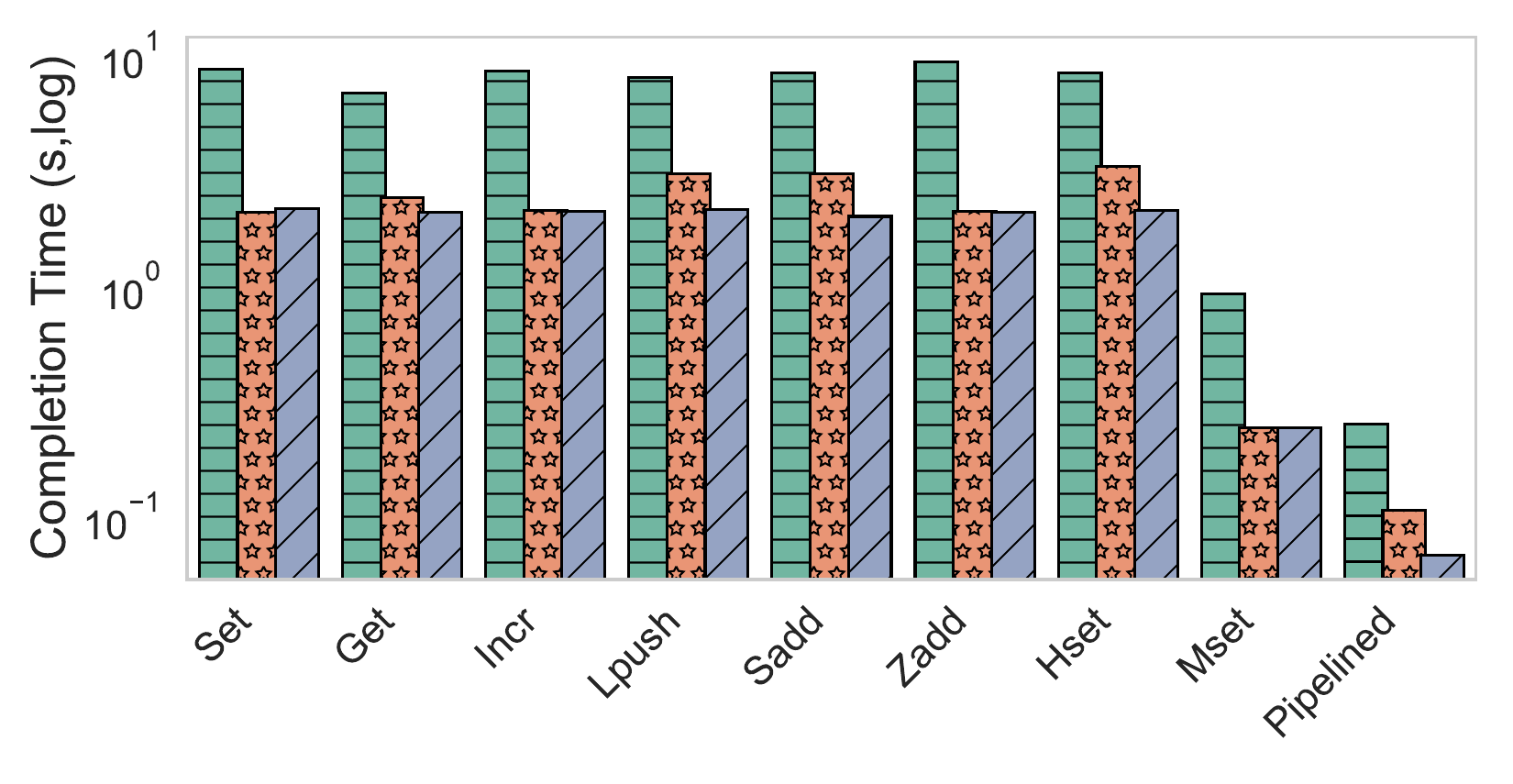}
    \label{fig:subfig1_redis_completion_time}
  \end{subfigure}%
  \hfill
  \begin{subfigure}[t]{0.49\textwidth}
    \centering
    \includegraphics[width=\linewidth]{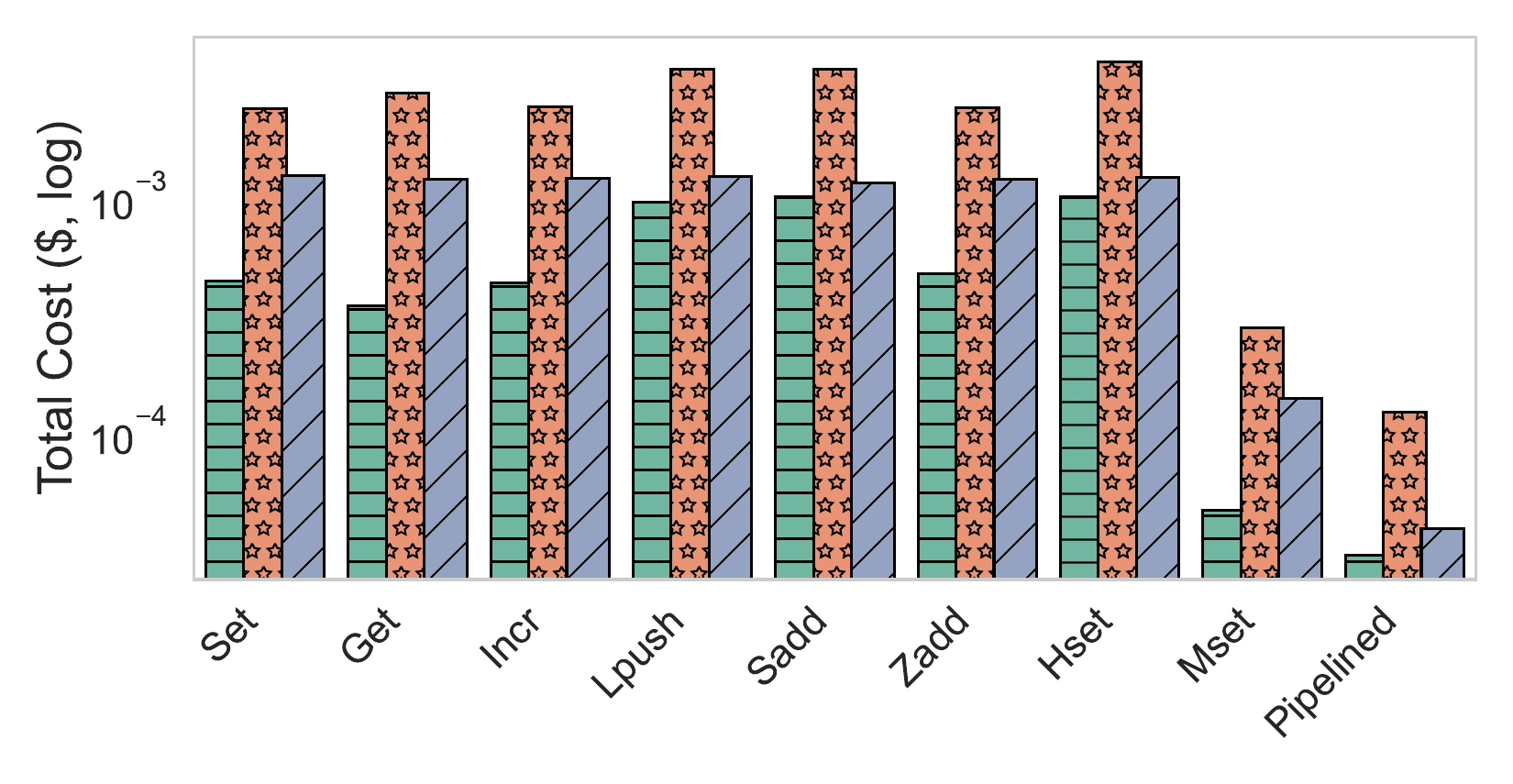}
    \label{fig:subfig1_redis_costs}
  \end{subfigure}%
  \hfill
  \vspace{-1.7em}
  \begin{subfigure}[t]{0.25\textwidth}
    \centering
    \includegraphics[width=\linewidth]{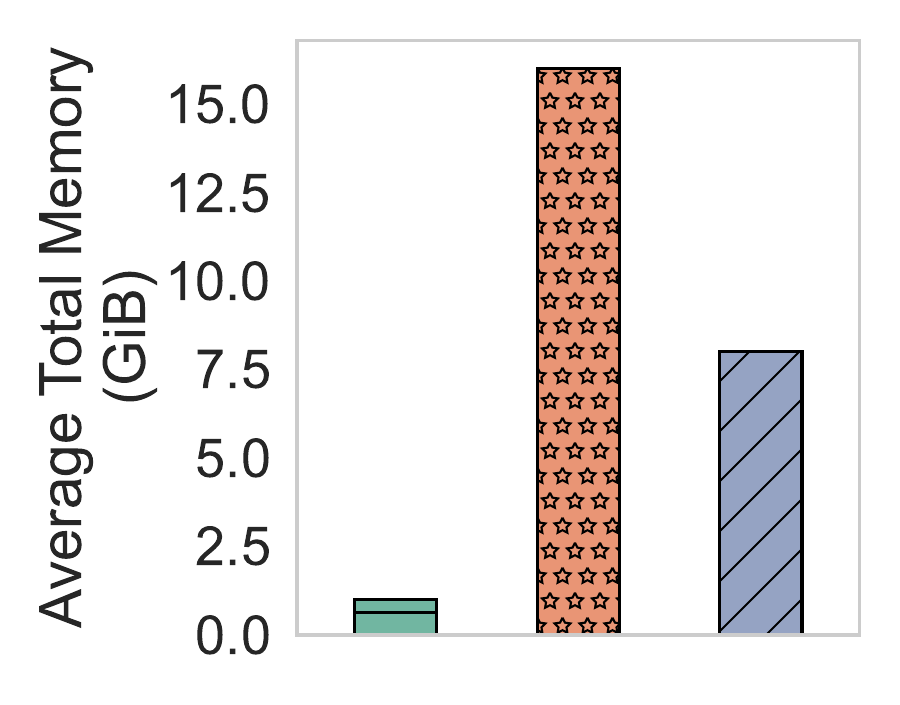}
    \label{fig:subfig2_redis_memory}
  \end{subfigure}%
  \begin{subfigure}[t]{0.25\textwidth}
    \centering
    \includegraphics[width=\linewidth]{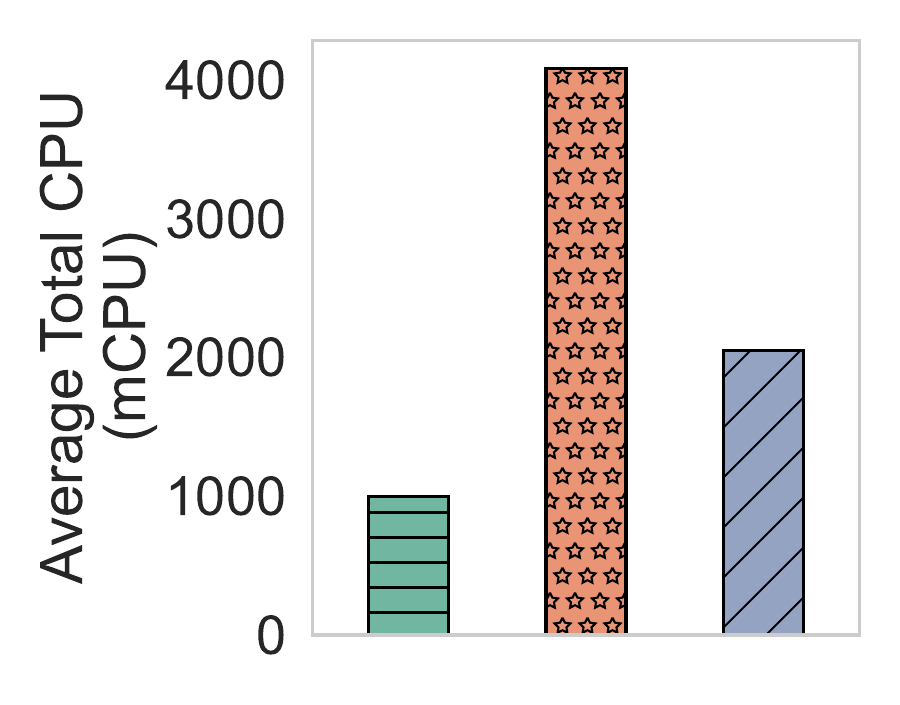}
    \label{fig:subfig3_redis_cpu}
  \end{subfigure}%
  \begin{subfigure}[t]{0.25\textwidth}
    \centering
    \includegraphics[width=\linewidth]{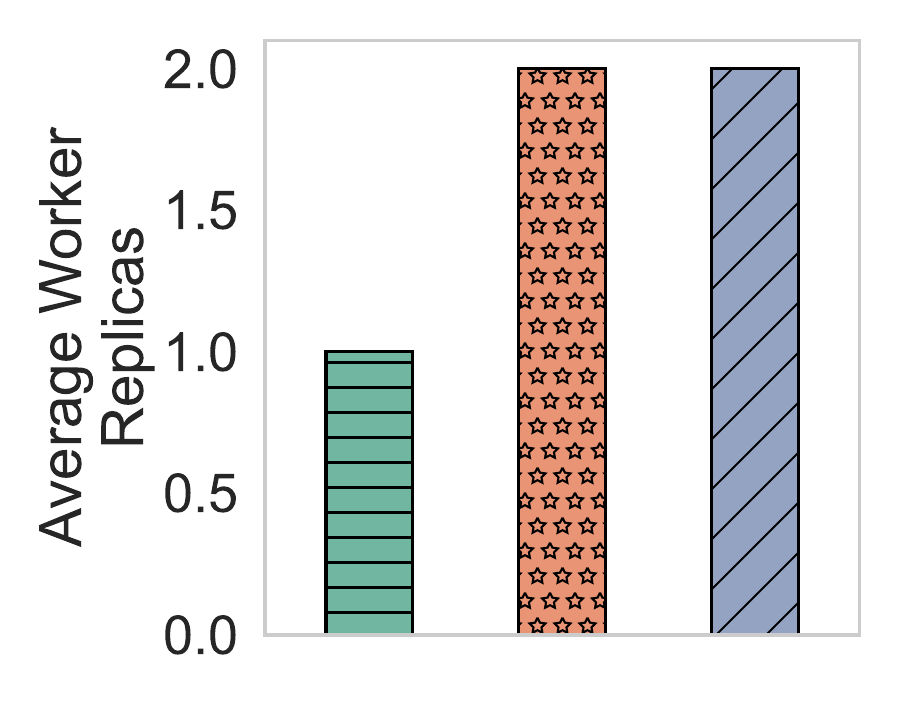}
    \label{fig:subfig4_redis_replicas}
  \end{subfigure}
  \vspace{-2em}
  \caption{Dynamic validation of Redis over 9 different benchmarks for different User Intents. Log-scale Completion times (top left) and Benchmark Processing Costs (top right) of benchmarks, allocations of Memory (bottom left), CPU as milliCPU, or $1/1000$ of a CPU (bottom mid), and replicas (bottom right).}
  \label{fig:redis_dynamic_validation}
  \vspace{-1em}
\end{figure*}

\section{Evaluation}
\label{sec:Evaluation}

\subsection{Experimental Setup}
\label{subsec: Experimental Setup}
We use a two-fold approach for evaluation using static and dynamic validation techniques borrowed from Software Engineering~\cite{static_dynamic_validation}. Our static validation benchmark consists of 180 data samples spread across 3 different difficulty levels of distributed systems. For each distributed system we have 6 different difficulty levels with 10 data samples per difficulty level forming 60 data samples per distributed system. For dynamic validation, we do actual deployment in a multi-tenant cloud environment using Kubernetes with the generated config by the LLM agent. After deployment, we run benchmark applications of each distributed system borrowed from their official documentation and continuously monitor several metrics such as benchmark completion time, cost, and resource consumption to evaluate performance alignment with the user intent and prompt. We use several different \emph{coding specialized LLMs} for evaluation for static validation including Qwen Coder 2.5-7B and Qwen Coder 2.5-14B~\cite{Qwen2.5-Coder}, DeepSeek Coder V2-16B~\cite{DeepSeek-Coder-V2}, CodeLlama-7B and CodeLlama-13B~\cite{CodeLlama}, and Llama 3.1-8B~\cite{Llama3_1}. The LLM agents are created using LangChain~\cite{Chase2022LangChain}, HuggingFace~\cite{HuggingFace}, and Ollama~\cite{Ollama} libraries in Python. For dynamic validation, we choose the best-performing model from these due to the steep cost overhead of dynamic validation-based evaluation. Notably, as per our knowledge, ours is the first work that does dynamic validation while related works only rely on static validation~\cite{kon2024iac-eval, li2024preconfig}.

\subsection{Results and Discussion}
\label{subsec: Results}

\subsubsection{Static Validation}
\label{subsubsec: Static Validation}
In this section, we evaluate the accuracy of different LLMs in generating valid configurations through static validation. The purpose of this experiment is to assess whether LLM-generated configurations align with user intent without requiring actual deployment. Table~\ref{tab:main_table} summarizes the results across three distributed systems: Dask, Redis, and Ray highlighting the best-performing optimization strategy for each model. We present the longer version of this table in the Appendix~\ref{sec:appendix}, which also includes the detailed level-wise results. 

\textbf{OB\ding{184} Among optimization techniques, FSL emerged as the most effective in improving configuration accuracy}. This aligns with findings from prior work, which highlight FSL’s ability to enhance structured generation by providing in-context learning examples~\cite{kon2024iac-eval, lian2024ciri}. The insights from our results in Table~\ref{tab:main_table} highlight a observation, \textbf{OB\ding{185} CoT used to enhance advanced reasoning abilities, did not result in noticeable improvements over vanilla models.} However, this technique can be useful to perform validation of configuration when static human-curated validators are not available which is why we have used it in LADs for this purpose.

\textbf{OB\ding{186} Relatively recent models despite their smaller size perform better than the larger but older models.} For example, Table~\ref{tab:main_table} shows Llama 3.1 (8B) and Qwen 2.5 (7B) both outperform CodeLlama (13B) and DeepSeek V2 (16B). This indicates that configuration generation tasks can be handled effectively by smaller, computationally efficient models rather than expensive large models. This discrepancy can be attributed to the newer architectures and improved training datasets of Llama 3.1 and Qwen2.5, compared to the older, fine-tuned CodeLlama models~\cite{qiao2024benchmarking}. Notably, CodeLlama is based on Llama 2, which is nearly two years old, suggesting that newer models benefit from higher-quality training data. In high-frequency API calling scenarios, this independence from large models is particularly beneficial, as it enables faster, more cost-effective automation without sacrificing accuracy. Analyzing model performance across different LLMs, Qwen2.5 14B was the overall best performer, consistently achieving the highest accuracy across multiple applications. Given its strong results, we select Qwen2.5 14B as the primary model for conducting all subsequent dynamic validations. For difficulty level-wise breakup of static validation results per distributed system please refer to Appendix~\ref{subsec: Appendix: Static Validation with LADs}.


\subsubsection{Dynamic Validation}
\label{subsubsec: Dynamic Validation}

The dynamic validation results in Figure~\ref{fig:redis_dynamic_validation} show the completion times and the processing costs of benchmarks for different benchmark applications run on Redis deployment via \proj. The Completion times and corresponding cost results reflect that \proj is taking user intent into account as ``Low Cost and Resource consumption'' leads to higher completion times due to less amount of resources allocated and this also leads to lower costs. Similarly, ``High scalability'' results in lesser completion time due to the abundant allocation of resources for executing the benchmark application but results in a higher cost. The more complex case of maintaining ``High Efficiency with balanced Scalability and Cost'' results in the lowest completion time for most benchmark applications and also balances between costs and resource consumption, costing lower than the ``High Scalability'' intent but higher than ``Low Cost and Resource consumption''. 

We observe a trend in memory, CPU, and replica resource consumption as shown in Figure~\ref{fig:redis_dynamic_validation}. The “Low Cost and Resource consumption” intent shows the lowest allocation, followed by “High Efficiency with balanced Scalability and Cost,” and finally “High scalability” exhibits the highest allocation. \emph{These results demonstrate that \proj can serve as a one-stop solution for accurate configuration generation and deployment of various distributed systems, aligning with both simple and complex user intents and dynamic, heterogeneous system resources.} For additional validation with other distributed systems, refer to Appendix~\ref{subsec: Appendix: Dynamic Validation with LADs}.
\begin{figure}
    \centering
    \includegraphics[width=0.8\linewidth]{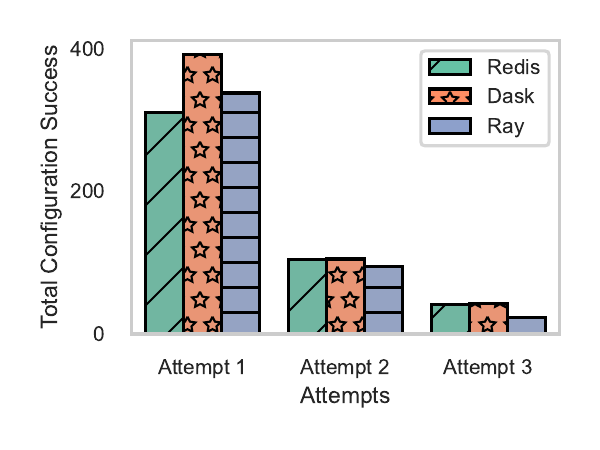}
    \vspace{-1.5em}
    \caption{Feedback-based prompt chaining}
    \vspace{-1.2em}
    \label{fig:Feedback_based_prompt_chaining}
\end{figure}
\paragraph{Feedback-based prompt chaining}
Figure~\ref{fig:Feedback_based_prompt_chaining} shows how feedback-based prompt chaining enables the resolution of errors and Warnings to get accurate configurations. \textbf{OB\ding{187} Approximately 94\% of configuration errors can be resolved through feedback-based prompt chaining technique in LADs by the third attempt.}

\begin{table}
  \centering
  \small
  \begin{tabular}{lcc}
    \noalign{\hrule height 1pt}
    \rowcolor{pink}
    \textbf{Systems} & \textbf{Avg. Latency (s)} & \textbf{Std. Dev. (s)} \\
    \hline
    Dask  & 9.282  & 0.849  \\
    Redis & 4.772  & 0.462  \\
    Ray   & 10.524 & 0.159  \\
    \noalign{\hrule height 1pt}
    
  \end{tabular}
  \caption{Overhead of Configuration Generation by via \proj per API call.}
  \label{tab:llm_overhead}
\end{table}

\begin{table}[h]
\centering
\small
\begin{tabular}{lcc}
\noalign{\hrule height 1pt}
\rowcolor{pink}
\textbf{Systems} & \textbf{Max Tokens} & \textbf{Cost per Config (\$)} \\
\hline
Dask   & 67  & 0.0000536 \\
Redis  & 183 & 0.0001464 \\
Ray    & 72  & 0.0000576 \\
\noalign{\hrule height 1pt}
\end{tabular}
\caption{Maximum tokens and cost per configuration for selected distributed systems.}
\vspace{-1em}
\label{tab:distributed_costs}
\end{table}

\subsection{Overhead of \proj}
\label{seubsec: Overhead of LADs}
As shown in Table~\ref{tab:llm_overhead} \proj takes between \emph{4 to 11 seconds} on average for configuration generation using LLM agents. This process is significantly more efficient than manually writing and tuning configurations to align with user intent in diverse and dynamically varying cloud environments. Table~\ref{tab:distributed_costs} shows the maximum tokens available for each distributed system configuration, along with their extremely low cost per configuration in the order of a \emph{few ten-thousandths of a dollar}. In contrast, skilled DevOps engineers charge high hourly rates, typically ranging from $\$40$ to $\$100$ per hour~\cite{UpworkDevOpsCost}. \textbf{OB\ding{188} By leveraging cost efficient LLMs with fewer than 20 Billion parameters for frequent API calls, \proj delivers rapid and affordable configuration generation. This makes it a particularly attractive solution for startups that do not have the budget to hire expert DevOps engineers but still want to avail themselves of these high-quality services.}
\section{Conclusion}
\label{sec: Conclusion}
In this work, we introduced LADs, a novel LLM-driven framework for automating cloud configuration and deployment, addressing the challenges of adaptability, efficiency, and robustness in AI-powered DevOps. By integrating techniques such as IP, RAG, FSL, CoT, and Feedback-Based Prompt Chaining, LADs not only generates accurate configurations but also provides maintenance based on real-time feedback. Our extensive evaluations demonstrate that LADs significantly reduces manual effort, optimizes resource utilization, and enhances system reliability, outperforming individual optimizations. Additionally, our benchmark and dataset provide a foundation for further research in automated DevOps validation. By open-sourcing LADs, we aim for innovations in automated cloud DevOps, enabling scalable, cost-effective, and error-resilient deployments.
\section*{Limitations}
\label{sec: Limitations}
\paragraph{Overhead for new Distributed Systems} Current limitations of our work include the one-time overhead of adding a distributed system to \proj. This process involves designing example shots and providing documentation and best practices. While this requires an initial manual effort, our future work includes automating this step by embedding web search ability in \proj to search for documentation and form shots with the help of agents.

\paragraph{\proj for unseen systems} Our current evaluation focuses on popular distributed systems that cater to different microservices, however, in the future we plan on extending \proj for unseen systems. This will allow \proj to serve as an automated DevOps solution for developers who create their own custom distributed systems unseen by state-of-the-art LLMs. This will require generating documentation and shots by parsing the code via LLM agents and is part of our future work. However, our current evaluation is focused on seen distributed systems as it is difficult to find unseen distributed systems that are also popular.

\section*{Ethical Considerations}
Our dataset includes only user intent, prompts, and Python validator functions. We did not collect any assets or any personally identifiable information. Our dynamic validation benchmark uses the publicly available benchmarks from official documentation of the distributed systems which are made public for evaluation of these systems. To the best of our knowledge, we did not violate any code of ethics with the experiments done in this paper. We reported technical details and results in the main paper and Appendix. Our system (LADs) has uses in real-world settings for automated DevOps. However, we caution that it must be used carefully, as the outputs are from an ML model and can have real-world consequences if used incorrectly.

\section*{Acknowledgments}
The work was supported in part by the NSF grants CSR-2106634 and CSR-2312785, the Amazon ML Systems Fellowship, the UMN GAGE Fellowship, and the Samsung Global Research Outreach Award.
\vspace{-0.7em}
\bibliography{main}

\clearpage
\onecolumn
\appendix

\section{Appendix}
\label{sec:appendix}
\subsection{Examples of optimizations in \proj}
\label{subsec: Examples of optimizations in LADs}
\begin{figure*}[htpb]
  \begin{center}
    \includegraphics[width=\textwidth]{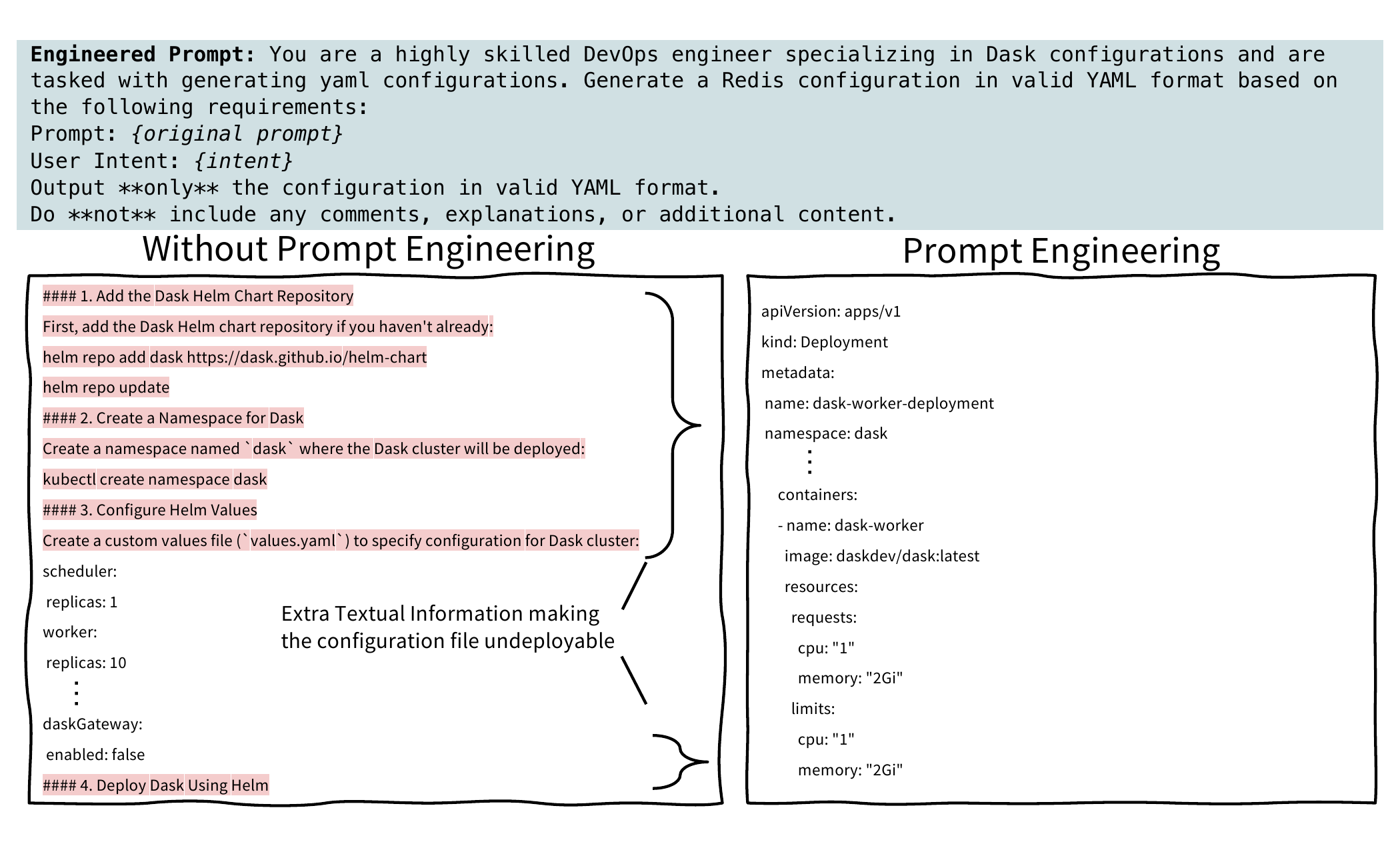}
    \vspace{-20pt}
    \caption{Example and impact of Instruction Prompting.}
    \label{fig:prompt_eng_example}
  \end{center}
  \vspace{-20pt}
\end{figure*}

\begin{figure}[t]
  \begin{center}
    \includegraphics[width=0.5\columnwidth]{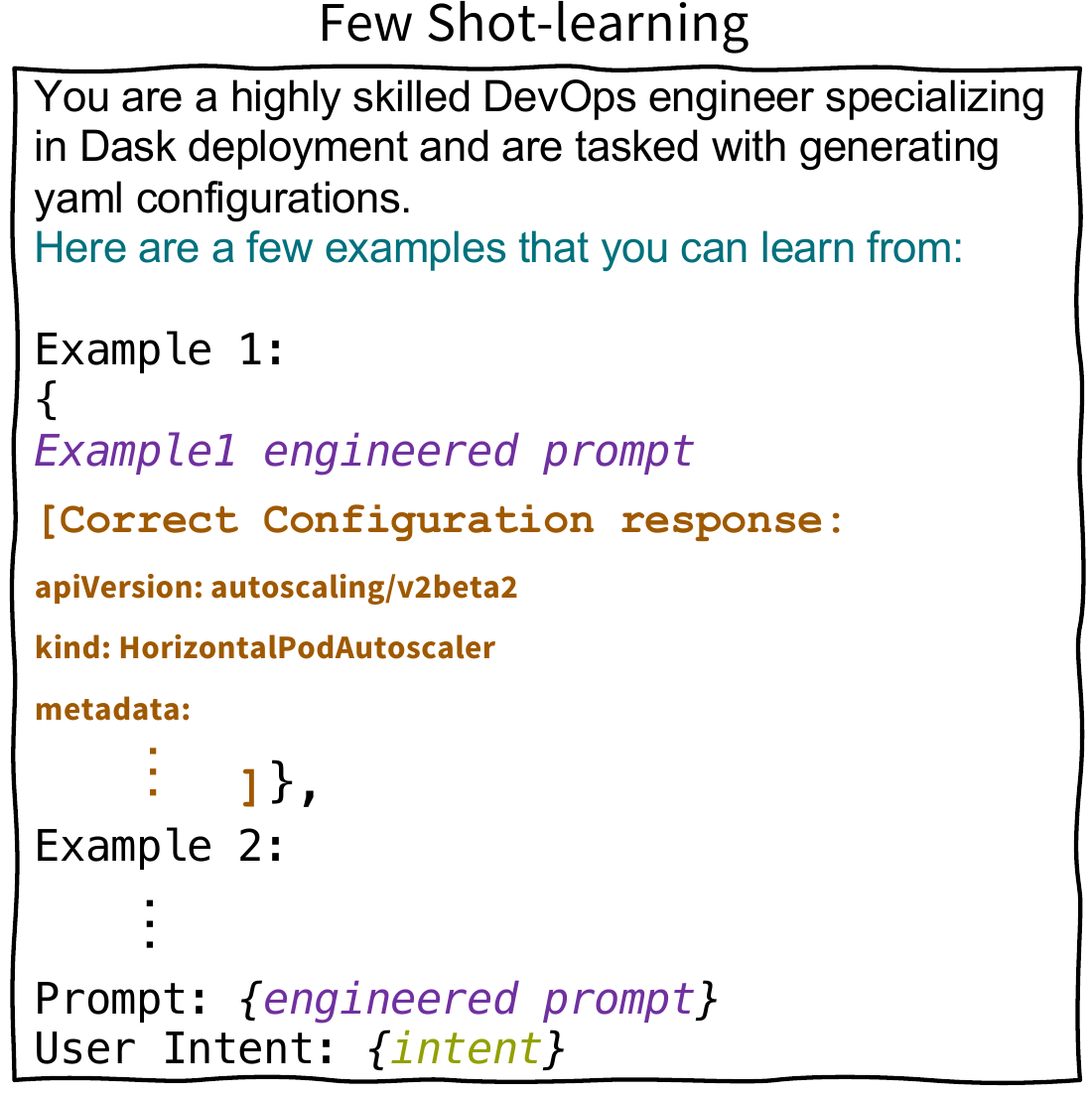}
    \caption{Example usage of few-shot technique for configuration generation in \proj.}
    \label{fig:few_shot_example}
  \end{center}
\end{figure}

In this section, we illustrate key optimization techniques employed in \proj, focusing on improving performance and efficiency in distributed systems through advanced prompting strategies. The following examples highlight how different prompting methodologies were utilized in the evaluations. Figure~\ref{fig:prompt_eng_example} showcases the application and impact of Instruction Prompting within \proj. Figure~\ref{fig:few_shot_example} demonstrates the use of the Few-Shot Learning (FSL) technique for automated configuration generation in \proj. These optimizations, among others, contribute to the overall effectiveness of \proj, enabling it to deliver enhanced automation and performance improvements across diverse distributed computing frameworks.

\subsection{Static Validation with \proj}
\label{subsec: Appendix: Static Validation with LADs}
In this section, we present the results of our static validation experiments conducted on three different distributed systems: Dask, Redis, and Ray. The performance metrics for these systems are detailed in Tables~\ref{table:Dask_Table}, \ref{table:Redis_Table}, and \ref{table:Ray_Table}.

To comprehensively evaluate the effectiveness of our approach, we tested our benchmark across six different LLMs using various prompting strategies, including Instruction Prompting, Chain-of-Thought (CoT), Retrieval-Augmented Generation (RAG), Few-Shot Learning (FSL), and our proposed \proj. The results consistently highlight that \proj outperformed all other optimization techniques across the distributed systems tested.

Among the LLMs evaluated, Qwen2.5-Coder-14B emerged as the best-performing model, demonstrating superior accuracy and efficiency across the benchmarks. This further underscores the effectiveness of integrating LADs with state-of-the-art language models to generate configurations for distributed system workflows.
\definecolor{pink}{HTML}{EAD5DC}
\definecolor{blue}{HTML}{EEE3E7}

\begin{table*}[t]
\centering
\small
\begin{tabular}{c|lccccccccc}
\noalign{\hrule height 1pt} 
\multirow{2}{*}{\footnotesize \textbf{\begin{tabular}[c]{@{}c@{}}Language\\ Model\end{tabular}}} & \multicolumn{1}{c|}{\multirow{2}{*}{\footnotesize\textbf{Optimizations}}} & \multicolumn{6}{c|}{\footnotesize\textit{\textbf{Levels}}} & \multirow{2}{*}{\footnotesize \textbf{\begin{tabular}[c]{@{}c@{}}Total\\ Accuracy\end{tabular}}} \\ \cline{3-8} 
 & \multicolumn{1}{c|}{} & \footnotesize\textbf{1 $(\uparrow)$} & \multicolumn{1}{c}{\footnotesize\textbf{2 $(\uparrow)$}} & \footnotesize\textbf{3 $(\uparrow)$} & \multicolumn{1}{c}{\footnotesize\textbf{4 $(\uparrow)$}} & \footnotesize\textbf{5 $(\uparrow)$} & \multicolumn{1}{c|}{\footnotesize\textbf{6 $(\uparrow)$}} \\

\hline

\multirow{5}{*}{\begin{tabular}[c]{@{}c@{}}\textbf{{CodeLlama 7b}}\end{tabular}}
 & \multicolumn{1}{l|}{\textbf{IP}} & 10 & 0 & 10 & 0 & 10 & \multicolumn{1}{l|}{0} &  5.00\\
 & \multicolumn{1}{l|}{\textbf{CoT}} & 0 & 0 & 0 & 0 & 0 & \multicolumn{1}{l|}{0} & 0.00\\
 & \multicolumn{1}{l|}{\textbf{RAG}} & 10 & 0 & 20 & 0 & 10 & \multicolumn{1}{l|}{0} & 6.67\\
 & \multicolumn{1}{l|}{\textbf{FSL}} & 40 & 0 & 100 & 0 & 30 & \multicolumn{1}{l|}{0} & 28.34\\
 & \multicolumn{1}{l|}{\cellcolor{pink!50}\textbf{LADs}} & \cellcolor{pink!50}40 & \cellcolor{pink!50}30 & \cellcolor{pink!50}100 & \cellcolor{pink!50}0 & \cellcolor{pink!50}30 & \multicolumn{1}{l|}{\cellcolor{pink!50}0} & \cellcolor{pink!50}\textbf{33.34}\\
\hline

\multirow{5}{*}{\begin{tabular}[c]{@{}c@{}}\textbf{{CodeLlama 13b}}\end{tabular}}
 & \multicolumn{1}{l|}{\textbf{IP}} & 10 & 0 & 10 & 0 & 10 & \multicolumn{1}{l|}{0} & 5.00\\
 & \multicolumn{1}{l|}{\textbf{CoT}} & 10 & 0 & 10 & 0 & 0 & \multicolumn{1}{l|}{0} & 3.34\\
 & \multicolumn{1}{l|}{\textbf{RAG}} & 20 & 10 & 10 & 0 & 10 & \multicolumn{1}{l|}{0} & 8.34\\
 & \multicolumn{1}{l|}{\textbf{FSL}} & 60 & 90 & 90 & 0 & 20 & \multicolumn{1}{l|}{0} & 43.34\\
 & \multicolumn{1}{l|}{\cellcolor{pink!50}\textbf{LADs}} & \cellcolor{pink!50}90 & \cellcolor{pink!50}80 & \cellcolor{pink!50}80 & \cellcolor{pink!50}10 & \cellcolor{pink!50}30 & \multicolumn{1}{l|}{\cellcolor{pink!50}0} & \cellcolor{pink!50}\textbf{48.34}\\
\hline

\multirow{5}{*}{\begin{tabular}[c]{@{}c@{}}\textbf{{Llama 3.1 8b}}\end{tabular}}
 & \multicolumn{1}{l|}{\textbf{IP}} & 30 & 0 & 40 & 0 & 40 & \multicolumn{1}{l|}{0} & 18.34\\
 & \multicolumn{1}{l|}{\textbf{CoT}} & 0 & 0 & 30 & 0 & 20 & \multicolumn{1}{l|}{0} & 8.34\\
 & \multicolumn{1}{l|}{\textbf{RAG}} & 40 & 0 & 40 & 0 & 30 & \multicolumn{1}{l|}{0} & 18.34\\
 & \multicolumn{1}{l|}{\textbf{FSL}} & 100 & 60 & 100 & 10 & 60 & \multicolumn{1}{l|}{0} & \textbf{46.67}\\
 & \multicolumn{1}{l|}{\cellcolor{pink!50}\textbf{LADs}} & \cellcolor{pink!50}90 & \cellcolor{pink!50}30 & \cellcolor{pink!50}100 & \cellcolor{pink!50}30 & \cellcolor{pink!50}50 & \multicolumn{1}{l|}{\cellcolor{pink!50}0} & \cellcolor{pink!50}40.00\\
\hline

\multirow{5}{*}{\begin{tabular}[c]{@{}c@{}}\textbf{{DeepSeekV2 16b}}\end{tabular}}
 & \multicolumn{1}{l|}{\textbf{IP}} & 10 & 0 & 10 & 0 & 30 & \multicolumn{1}{l|}{0} & 8.34\\
 & \multicolumn{1}{l|}{\textbf{CoT}} & 20 & 30 & 30 & 0 & 40 & \multicolumn{1}{l|}{0} & 20.00\\
 & \multicolumn{1}{l|}{\textbf{RAG}} & 10 & 10 & 20 & 0 & 20 & \multicolumn{1}{l|}{0} & 10.00\\
 & \multicolumn{1}{l|}{\textbf{FSL}} & 80 & 80 & 90 & 0 & 40 & \multicolumn{1}{l|}{0} & 48.33\\
 & \multicolumn{1}{l|}{\cellcolor{pink!50}\textbf{LADs}} & \cellcolor{pink!50}80 & \cellcolor{pink!50}80 & \cellcolor{pink!50}100 & \cellcolor{pink!50}0 & \cellcolor{pink!50}40 & \multicolumn{1}{l|}{\cellcolor{pink!50}0} & \cellcolor{pink!50}\textbf{50.00}\\
\hline

\multirow{5}{*}{\begin{tabular}[c]{@{}c@{}}\textbf{{Qwen2.5 7b}}\end{tabular}}
 & \multicolumn{1}{l|}{\textbf{IP}} & 30 & 30 & 30 & 0 & 40 & \multicolumn{1}{l|}{0} & 21.67\\
 & \multicolumn{1}{l|}{\textbf{CoT}} & 40 & 0 & 30 & 0 & 30 & \multicolumn{1}{l|}{0} & 16.67\\
 & \multicolumn{1}{l|}{\textbf{RAG}} & 20 & 30 & 10 & 0 & 30 & \multicolumn{1}{l|}{0} & 15.00\\
 & \multicolumn{1}{l|}{\textbf{FSL}} & 90 & 90 & 100 & 0 & 50 & \multicolumn{1}{l|}{0} & 55.00\\
 & \multicolumn{1}{l|}{\cellcolor{pink!50}\textbf{LADs}} & \cellcolor{pink!50}100 & \cellcolor{pink!50}90 & \cellcolor{pink!50}100 & \cellcolor{pink!50}0 & \cellcolor{pink!50}50 & \multicolumn{1}{l|}{\cellcolor{pink!50}0} & \cellcolor{pink!50}\textbf{56.67}\\
\hline

\multirow{5}{*}{\begin{tabular}[c]{@{}c@{}}\textbf{{Qwen2.5 14b}}\end{tabular}}
 & \multicolumn{1}{l|}{\textbf{IP}} & 90 & 60 & 90 & 10 & 40 & \multicolumn{1}{l|}{0} & 48.34\\
 & \multicolumn{1}{l|}{\textbf{CoT}} & 80 & 50 & 70 & 20 & 50 & \multicolumn{1}{l|}{0} & 45.00\\
 & \multicolumn{1}{l|}{\textbf{RAG}} & 20 & 30 & 10 & 0 & 30 & \multicolumn{1}{l|}{20} & 18.34\\
 & \multicolumn{1}{l|}{\textbf{FSL}} & 100 & 90 & 100 & 0 & 50 & \multicolumn{1}{l|}{0} & 56.67\\
 & \multicolumn{1}{l|}{\cellcolor{pink!50}\textbf{LADs}} & \cellcolor{pink!50}100 & \cellcolor{pink!50}90 & \cellcolor{pink!50}100 & \cellcolor{pink!50}40 & \cellcolor{pink!50}50 & \multicolumn{1}{l|}{\cellcolor{pink!50}30} & \cellcolor{pink!50}\textbf{\underline{68.34}}\\


\noalign{\hrule height 1pt}
\end{tabular}
\caption{\textbf{Performance of Various LLMs on Dask Application with Different Language Models and Optimizations.}
This table presents the accuracy of different large language models (LLMs) on a Dask application under various optimization strategies across multiple difficulty levels. The optimizations include IP, CoT, RAG, FSL, and LADs. The highest-performing optimization for each LLM is highlighted in bold, while the overall best-performing LLM-optimization pair is highlighted in bold and underlined.}
\label{table:Dask_Table}
\end{table*}
\begin{table*}[t]
\centering
\small
\begin{tabular}{c|lccccccccc}
\noalign{\hrule height 1pt} 
\multirow{2}{*}{\footnotesize \textbf{\begin{tabular}[c]{@{}c@{}}Language\\ Model\end{tabular}}} & \multicolumn{1}{c|}{\multirow{2}{*}{\footnotesize\textbf{Optimizations}}} & \multicolumn{6}{c|}{\footnotesize\textit{\textbf{Levels}}} & \multirow{2}{*}{\footnotesize \textbf{\begin{tabular}[c]{@{}c@{}}Total\\ Accuracy\end{tabular}}} \\ \cline{3-8} 
 & \multicolumn{1}{c|}{} & \footnotesize\textbf{1 $(\uparrow)$} & \multicolumn{1}{c}{\footnotesize\textbf{2 $(\uparrow)$}} & \footnotesize\textbf{3 $(\uparrow)$} & \multicolumn{1}{c}{\footnotesize\textbf{4 $(\uparrow)$}} & \footnotesize\textbf{5 $(\uparrow)$} & \multicolumn{1}{c|}{\footnotesize\textbf{6 $(\uparrow)$}} \\

\hline

\multirow{5}{*}{\begin{tabular}[c]{@{}c@{}}\textbf{{CodeLlama 7b}}\end{tabular}}
 & \multicolumn{1}{l|}{\textbf{IP}} & 30 & 10 & 30 & 10 & 20 & \multicolumn{1}{l|}{0} &  16.67\\
 & \multicolumn{1}{l|}{\textbf{CoT}} & 0 & 0 & 0 & 0 & 10 & \multicolumn{1}{l|}{0} & 1.67\\
 & \multicolumn{1}{l|}{\textbf{RAG}} & 30 & 0 & 20 & 30 & 20 & \multicolumn{1}{l|}{0} & 16.67\\
 & \multicolumn{1}{l|}{\textbf{FSL}} & 40 & 80 & 10 & 10 & 20 & \multicolumn{1}{l|}{0} & 26.67\\
 & \multicolumn{1}{l|}{\cellcolor{pink!50}\textbf{LADs}} & \cellcolor{pink!50}40 & \cellcolor{pink!50}80 & \cellcolor{pink!50}20 & \cellcolor{pink!50}20 & \cellcolor{pink!50}10 & \multicolumn{1}{l|}{\cellcolor{pink!50}0} & \cellcolor{pink!50}\textbf{28.34}\\
\hline

\multirow{5}{*}{\begin{tabular}[c]{@{}c@{}}\textbf{{CodeLlama 13b}}\end{tabular}}
 & \multicolumn{1}{l|}{\textbf{IP}} & 40 & 10 & 10 & 10 & 20 & \multicolumn{1}{l|}{0} & 15.00\\
 & \multicolumn{1}{l|}{\textbf{CoT}} & 10 & 0 & 0 & 0 & 0 & \multicolumn{1}{l|}{0} & 1.67\\
 & \multicolumn{1}{l|}{\textbf{RAG}} & 20 & 10 & 10 & 10 & 30 & \multicolumn{1}{l|}{0} & 13.33\\
 & \multicolumn{1}{l|}{\textbf{FSL}} & 50 & 60 & 10 & 10 & 20 & \multicolumn{1}{l|}{0} & \textbf{25.00}\\
 & \multicolumn{1}{l|}{\cellcolor{pink!50}\textbf{LADs}} & \cellcolor{pink!50}50 & \cellcolor{pink!50}50 & \cellcolor{pink!50}10 & \cellcolor{pink!50}10 & \cellcolor{pink!50}20 & \multicolumn{1}{l|}{\cellcolor{pink!50}0} & \cellcolor{pink!50}23.34\\
\hline

\multirow{5}{*}{\begin{tabular}[c]{@{}c@{}}\textbf{{Llama 3.1 8b}}\end{tabular}}
 & \multicolumn{1}{l|}{\textbf{IP}} & 40 & 10 & 40 & 20 & 20 & \multicolumn{1}{l|}{0} & 21.67\\
 & \multicolumn{1}{l|}{\textbf{CoT}} & 40 & 20 & 40 & 20 & 20 & \multicolumn{1}{l|}{0} & 23.34\\
 & \multicolumn{1}{l|}{\textbf{RAG}} & 40 & 10 & 40 & 20 & 20 & \multicolumn{1}{l|}{0} & 21.67\\
 & \multicolumn{1}{l|}{\textbf{FSL}} & 60 & 90 & 30 & 20 & 10 & \multicolumn{1}{l|}{0} & 35.00\\
 & \multicolumn{1}{l|}{\cellcolor{pink!50}\textbf{LADs}} & \cellcolor{pink!50}40 & \cellcolor{pink!50}100 & \cellcolor{pink!50}50 & \cellcolor{pink!50}20 & \cellcolor{pink!50}10 & \multicolumn{1}{l|}{\cellcolor{pink!50}0} & \cellcolor{pink!50}\textbf{36.67}\\
\hline

\multirow{5}{*}{\begin{tabular}[c]{@{}c@{}}\textbf{{DeepSeekV2 16b}}\end{tabular}}
 & \multicolumn{1}{l|}{\textbf{IP}} & 40 & 20 & 30 & 10 & 20 & \multicolumn{1}{l|}{0} & 20.00\\
 & \multicolumn{1}{l|}{\textbf{CoT}} & 40 & 20 & 40 & 20 & 30 & \multicolumn{1}{l|}{0} & 25.00\\
 & \multicolumn{1}{l|}{\textbf{RAG}} & 40 & 10 & 40 & 20 & 20 & \multicolumn{1}{l|}{10} & 23.34\\
 & \multicolumn{1}{l|}{\textbf{FSL}} & 80 & 40 & 30 & 10 & 20 & \multicolumn{1}{l|}{0} & 30.00\\
 & \multicolumn{1}{l|}{\cellcolor{pink!50}\textbf{LADs}} & \cellcolor{pink!50}80 & \cellcolor{pink!50}90 & \cellcolor{pink!50}30 & \cellcolor{pink!50}10 & \cellcolor{pink!50}20 & \multicolumn{1}{l|}{\cellcolor{pink!50}0} & \cellcolor{pink!50}\textbf{38.34}\\
\hline

\multirow{5}{*}{\begin{tabular}[c]{@{}c@{}}\textbf{{Qwen2.5 7b}}\end{tabular}}
 & \multicolumn{1}{l|}{\textbf{IP}} & 50 & 10 & 20 & 10 & 30 & \multicolumn{1}{l|}{0} & 20.00\\
 & \multicolumn{1}{l|}{\textbf{CoT}} & 40 & 10 & 40 & 10 & 20 & \multicolumn{1}{l|}{0} & 20.00\\
 & \multicolumn{1}{l|}{\textbf{RAG}} & 30 & 0 & 20 & 10 & 30 & \multicolumn{1}{l|}{10} & 16.67\\
 & \multicolumn{1}{l|}{\textbf{FSL}} & 70 &80 & 10 & 10 & 10 & \multicolumn{1}{l|}{0} & 30.00\\
 & \multicolumn{1}{l|}{\cellcolor{pink!50}\textbf{LADs}} & \cellcolor{pink!50}80 & \cellcolor{pink!50}80 & \cellcolor{pink!50}10 & \cellcolor{pink!50}10 & \cellcolor{pink!50}10 & \multicolumn{1}{l|}{\cellcolor{pink!50}0} & \cellcolor{pink!50}\textbf{31.67}\\
\hline

\multirow{5}{*}{\begin{tabular}[c]{@{}c@{}}\textbf{{Qwen2.5 14b}}\end{tabular}}
 & \multicolumn{1}{l|}{\textbf{IP}} & 90 & 30 & 70 & 50 & 20 & \multicolumn{1}{l|}{10} & 45.00\\
 & \multicolumn{1}{l|}{\textbf{CoT}} & 90 & 20 & 30 & 60 & 40 & \multicolumn{1}{l|}{20} & 43.34\\
 & \multicolumn{1}{l|}{\textbf{RAG}} & 80 & 30 & 50 & 0 & 30 & \multicolumn{1}{l|}{0} & 31.67\\
 & \multicolumn{1}{l|}{\textbf{FSL}} & 80 & 40 & 20 & 10 & 20 & \multicolumn{1}{l|}{20} & 31.67\\
 & \multicolumn{1}{l|}{\cellcolor{pink!50}\textbf{LADs}} & \cellcolor{pink!50}90 & \cellcolor{pink!50}100 & \cellcolor{pink!50}60 & \cellcolor{pink!50}50 & \cellcolor{pink!50}30 & \multicolumn{1}{l|}{\cellcolor{pink!50}20} & \cellcolor{pink!50}\textbf{\underline{58.33}}\\

\noalign{\hrule height 1pt}
\end{tabular}
\caption{\textbf{Performance of Various LLMs on Redis Application with Different Language Models and Optimizations.}
This table presents the accuracy of different large language models (LLMs) on a Redis application under various optimization strategies across multiple difficulty levels. The optimizations include IP, CoT, RAG, FSL, and LADs. The highest-performing optimization for each LLM is highlighted in bold, while the overall best-performing LLM-optimization pair is highlighted in bold and underlined.}
\label{table:Redis_Table}
\end{table*}
\begin{table*}[t]
\centering
\small
\begin{tabular}{c|lccccccccc}
\noalign{\hrule height 1pt} 
\multirow{2}{*}{\footnotesize \textbf{\begin{tabular}[c]{@{}c@{}}Language\\ Model\end{tabular}}} & \multicolumn{1}{c|}{\multirow{2}{*}{\footnotesize\textbf{Optimizations}}} & \multicolumn{6}{c|}{\footnotesize\textit{\textbf{Levels}}} & \multirow{2}{*}{\footnotesize \textbf{\begin{tabular}[c]{@{}c@{}}Total\\ Accuracy\end{tabular}}} \\ \cline{3-8} 
 & \multicolumn{1}{c|}{} & \footnotesize\textbf{1 $(\uparrow)$} & \multicolumn{1}{c}{\footnotesize\textbf{2 $(\uparrow)$}} & \footnotesize\textbf{3 $(\uparrow)$} & \multicolumn{1}{c}{\footnotesize\textbf{4 $(\uparrow)$}} & \footnotesize\textbf{5 $(\uparrow)$} & \multicolumn{1}{c|}{\footnotesize\textbf{6 $(\uparrow)$}} \\

\hline

\multirow{5}{*}{\begin{tabular}[c]{@{}c@{}}\textbf{{CodeLlama 7b}}\end{tabular}}
 & \multicolumn{1}{l|}{\textbf{IP}} & 10 & 0& 10 & 0& 0 & \multicolumn{1}{l|}{0} & 3.34\\
 & \multicolumn{1}{l|}{\textbf{CoT}} & 0 & 0 & 0 & 0 & 0 & \multicolumn{1}{l|}{0} & 0.00\\
 & \multicolumn{1}{l|}{\textbf{RAG}} & 0 & 0 & 10 & 0 & 0 & \multicolumn{1}{l|}{0} & 1.67\\
 & \multicolumn{1}{l|}{\textbf{FSL}} & 60 & 40 & 50 & 30 & 0 & \multicolumn{1}{l|}{30} & 35.00\\
 & \multicolumn{1}{l|}{\cellcolor{pink!50}\textbf{Combinations}} & \cellcolor{pink!50}60 & \cellcolor{pink!50}50 & \cellcolor{pink!50}90 & \cellcolor{pink!50}40 & \cellcolor{pink!50}0 & \multicolumn{1}{l|}{\cellcolor{pink!50}10} & \cellcolor{pink!50}\textbf{41.67}\\
\hline

\multirow{5}{*}{\begin{tabular}[c]{@{}c@{}}\textbf{{CodeLlama 13b}}\end{tabular}}
 & \multicolumn{1}{l|}{\textbf{IP}} & 0 & 0 & 0 & 0 & 0 & \multicolumn{1}{l|}{0} & 0.00\\
 & \multicolumn{1}{l|}{\textbf{CoT}} & 0 & 0 & 10 & 0 & 0 & \multicolumn{1}{l|}{0} & 1.67\\
 & \multicolumn{1}{l|}{\textbf{RAG}} & 0 & 0 & 0 & 0 & 0 & \multicolumn{1}{l|}{0} & 0.00\\
 & \multicolumn{1}{l|}{\textbf{FSL}} & 60 & 40	& 50 & 30 & 0 & \multicolumn{1}{l|}{30} & 35.00\\
 & \multicolumn{1}{l|}{\cellcolor{pink!50}\textbf{LADs}} & \cellcolor{pink!50}60 & \cellcolor{pink!50}50 & \cellcolor{pink!50}90 & \cellcolor{pink!50}40 & \cellcolor{pink!50}0 & \multicolumn{1}{l|}{\cellcolor{pink!50}10} & \cellcolor{pink!50}\textbf{41.67}\\
\hline

\multirow{5}{*}{\begin{tabular}[c]{@{}c@{}}\textbf{{Llama 3.1 8b}}\end{tabular}}
 & \multicolumn{1}{l|}{\textbf{IP}} & 10 & 0 & 10 & 0 & 0 & \multicolumn{1}{l|}{0} & 3.34\\
 & \multicolumn{1}{l|}{\textbf{CoT}} & 10 & 0 & 10 & 0 & 0 & \multicolumn{1}{l|}{0} & 3.34\\
 & \multicolumn{1}{l|}{\textbf{RAG}} & 10 & 0 & 10 & 0 & 0 & \multicolumn{1}{l|}{0} & 3.34\\
 & \multicolumn{1}{l|}{\textbf{FSL}} & 0 & 60 & 100 & 80 & 30 & \multicolumn{1}{l|}{40} & 51.67\\
 & \multicolumn{1}{l|}{\cellcolor{pink!50}\textbf{LADs}} & \cellcolor{pink!50}60 & \cellcolor{pink!50}40 & \cellcolor{pink!50}100 & \cellcolor{pink!50}70 & \cellcolor{pink!50}30 & \multicolumn{1}{l|}{\cellcolor{pink!50}40} & \cellcolor{pink!50}\textbf{56.67}\\
\hline

\multirow{5}{*}{\begin{tabular}[c]{@{}c@{}}\textbf{{DeepSeekV2 16b}}\end{tabular}}
 & \multicolumn{1}{l|}{\textbf{IP}} & 20 & 0 & 10 & 0 & 0 & \multicolumn{1}{l|}{0} & 5.00\\
 & \multicolumn{1}{l|}{\textbf{CoT}} & 10 & 0 & 20 & 0 & 0 & \multicolumn{1}{l|}{0} & 5.00\\
 & \multicolumn{1}{l|}{\textbf{RAG}} & 30 & 40 & 10 & 0 & 0 & \multicolumn{1}{l|}{0} & 13.34\\
 & \multicolumn{1}{l|}{\textbf{FSL}} & 60 & 60 & 50 & 0 & 0 & \multicolumn{1}{l|}{0} & 28.34\\
 & \multicolumn{1}{l|}{\cellcolor{pink!50}\textbf{LADs}} & \cellcolor{pink!50}70 & \cellcolor{pink!50}60 & \cellcolor{pink!50}50 & \cellcolor{pink!50}0 & \cellcolor{pink!50}0 & \multicolumn{1}{l|}{\cellcolor{pink!50}0} & \cellcolor{pink!50}\textbf{30.00}\\
\hline

\multirow{5}{*}{\begin{tabular}[c]{@{}c@{}}\textbf{{Qwen2.5 7b}}\end{tabular}}
 & \multicolumn{1}{l|}{\textbf{IP}} & 60 & 40 & 20 & 0 & 0 & \multicolumn{1}{l|}{0} & 20.00\\
 & \multicolumn{1}{l|}{\textbf{CoT}} & 70 & 40 & 10 & 0 & 0 & \multicolumn{1}{l|}{0} & 20.00\\
 & \multicolumn{1}{l|}{\textbf{RAG}} & 50 & 30 & 30 & 0 & 0 & \multicolumn{1}{l|}{0} & 18.34\\
 & \multicolumn{1}{l|}{\textbf{FSL}} & 80 & 30 & 90 & 60 & 60 & \multicolumn{1}{l|}{50} & \textbf{61.67}\\
 & \multicolumn{1}{l|}{\cellcolor{pink!50}\textbf{LADs}} & \cellcolor{pink!50}60 & \cellcolor{pink!50}40 & \cellcolor{pink!50}100 & \cellcolor{pink!50}70 & \cellcolor{pink!50}20 & \multicolumn{1}{l|}{\cellcolor{pink!50}30} & \cellcolor{pink!50}53.34\\
\hline

\multirow{5}{*}{\begin{tabular}[c]{@{}c@{}}\textbf{{Qwen2.5 14b}}\end{tabular}}
 & \multicolumn{1}{l|}{\textbf{IP}} & 80 & 50 & 50 & 0 & 0 & \multicolumn{1}{l|}{0} & 30.00\\
 & \multicolumn{1}{l|}{\textbf{CoT}} & 90 & 50 & 20 & 0 & 0 & \multicolumn{1}{l|}{0} & 26.67\\
 & \multicolumn{1}{l|}{\textbf{RAG}} & 50 & 30 & 30 & 0 & 0 & \multicolumn{1}{l|}{0} & 18.34\\
 & \multicolumn{1}{l|}{\textbf{FSL}} & 100 & 70 & 100 & 50 & 60 & \multicolumn{1}{l|}{0} & 63.34\\
 & \multicolumn{1}{l|}{\cellcolor{pink!50}\textbf{LADs}} & \cellcolor{pink!50}50 & \cellcolor{pink!50}70 & \cellcolor{pink!50}100 & \cellcolor{pink!50}70 & \cellcolor{pink!50}70 & \multicolumn{1}{l|}{\cellcolor{pink!50}60} & \cellcolor{pink!50}\textbf{\underline{70.00}}\\


\noalign{\hrule height 1pt}
\end{tabular}
\caption{\textbf{Performance of Various LLMs on Ray Application with Different Language Models and Optimizations.}
This table presents the accuracy of different large language models (LLMs) on a Ray application under various optimization strategies across multiple difficulty levels. The optimizations include IP, CoT, RAG, FSL, and LADs. The highest-performing optimization for each LLM is highlighted in bold, while the overall best-performing LLM-optimization pair is highlighted in bold and underlined.}
\label{table:Ray_Table}
\end{table*}

\subsection{Dynamic Validation with \proj}
\label{subsec: Appendix: Dynamic Validation with LADs}

\begin{figure*}[ht]
  \centering
    \includegraphics[width=0.8\linewidth]{figures/legend_aws_lambda_time_comparison_dask.pdf}
    \hfill
  \begin{subfigure}[t]{0.49\textwidth}
    \centering
    \includegraphics[width=\linewidth]{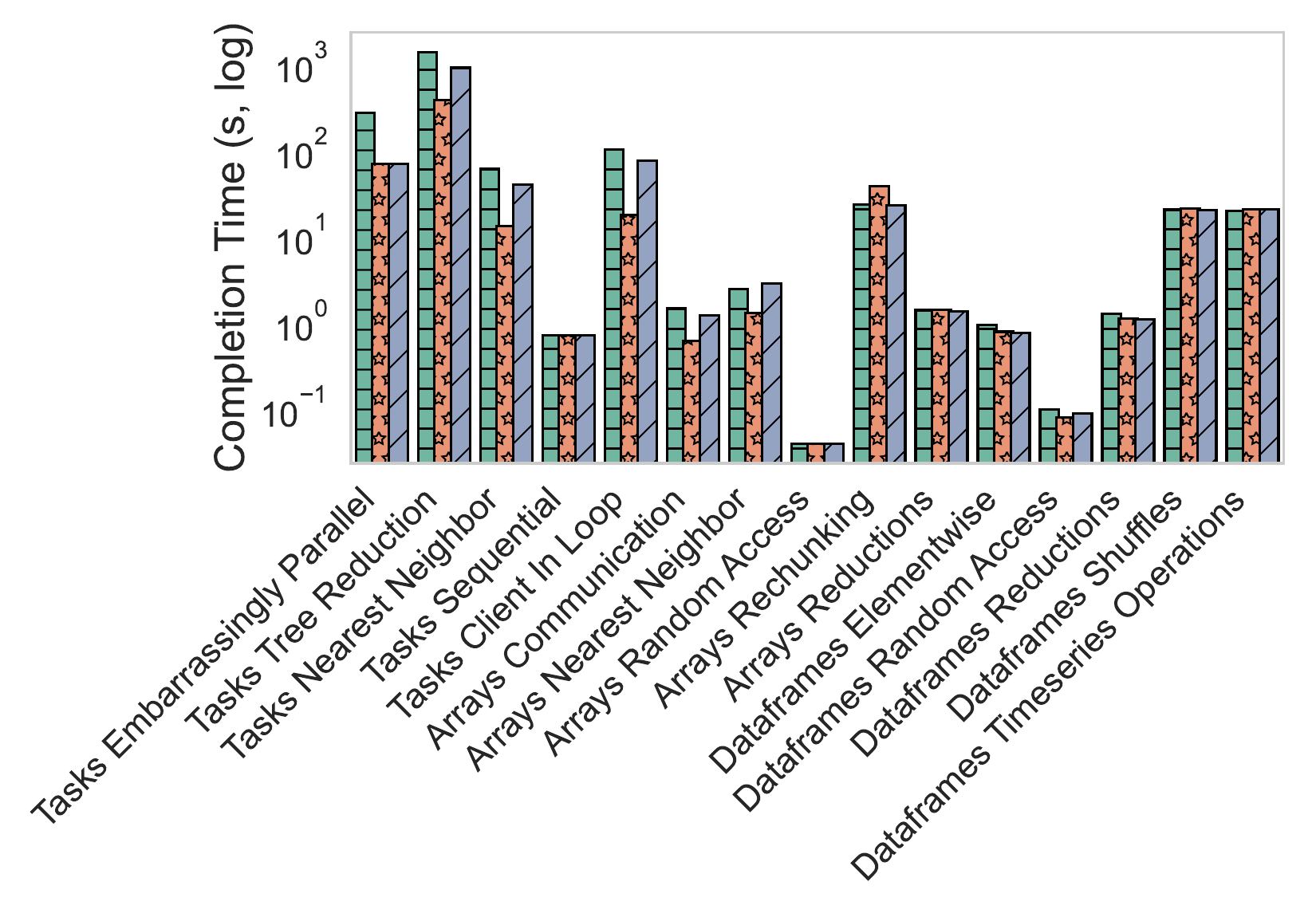}
    \label{fig:subfig1_dask_completion_time}
  \end{subfigure}%
  \hfill
      \begin{subfigure}[t]{0.49\textwidth}
    \centering
    \includegraphics[width=\linewidth]{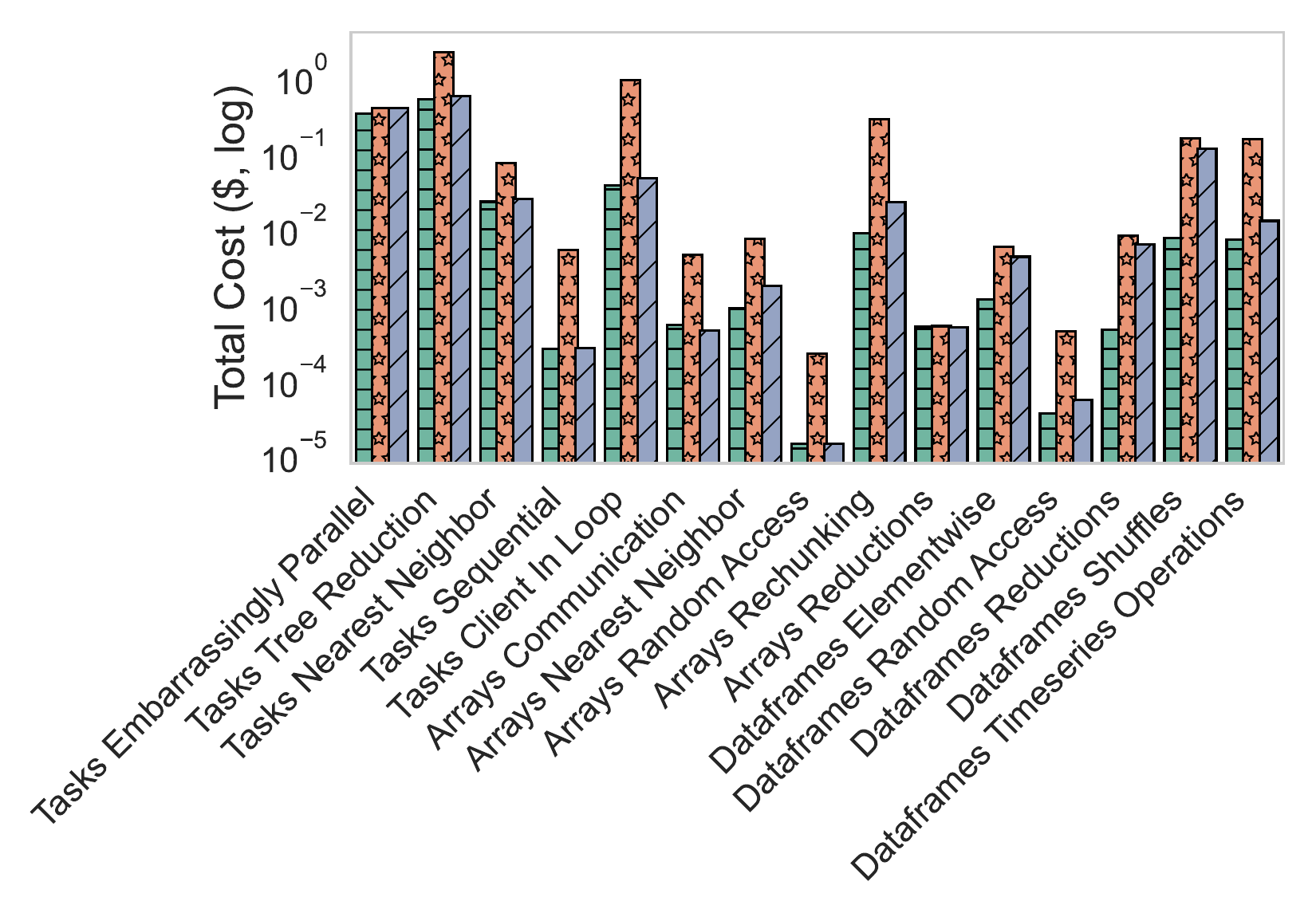}
    \label{fig:subfig1_dask_costs}
  \end{subfigure}%
  \hfill
  \begin{subfigure}[t]{0.25\textwidth}
    \centering
    \includegraphics[width=\linewidth]{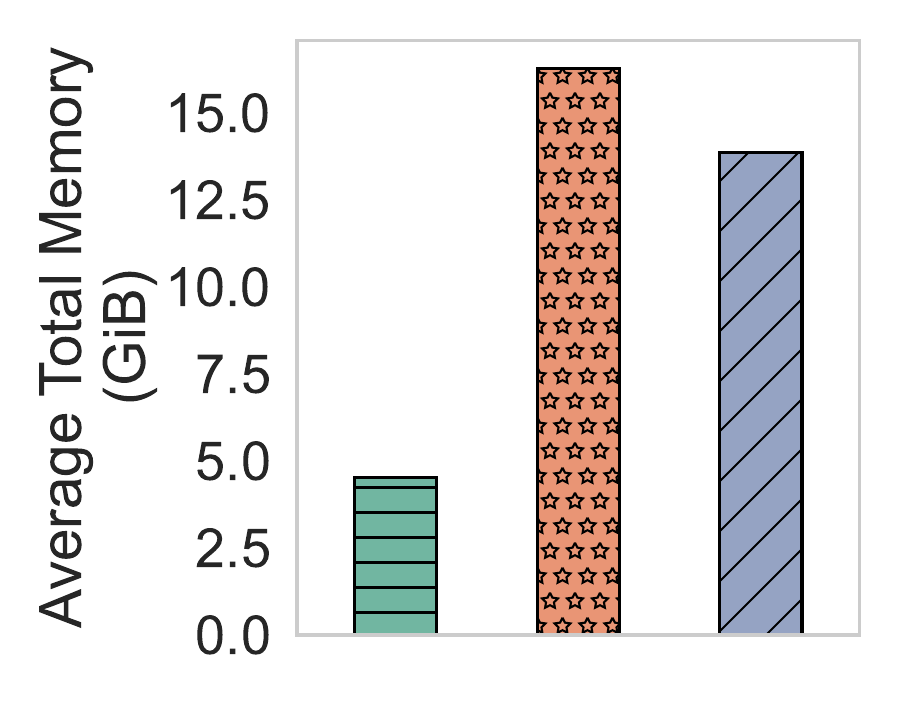}
    \label{fig:subfig2_dask_memory}
  \end{subfigure}%
  \begin{subfigure}[t]{0.25\textwidth}
    \centering
    \includegraphics[width=\linewidth]{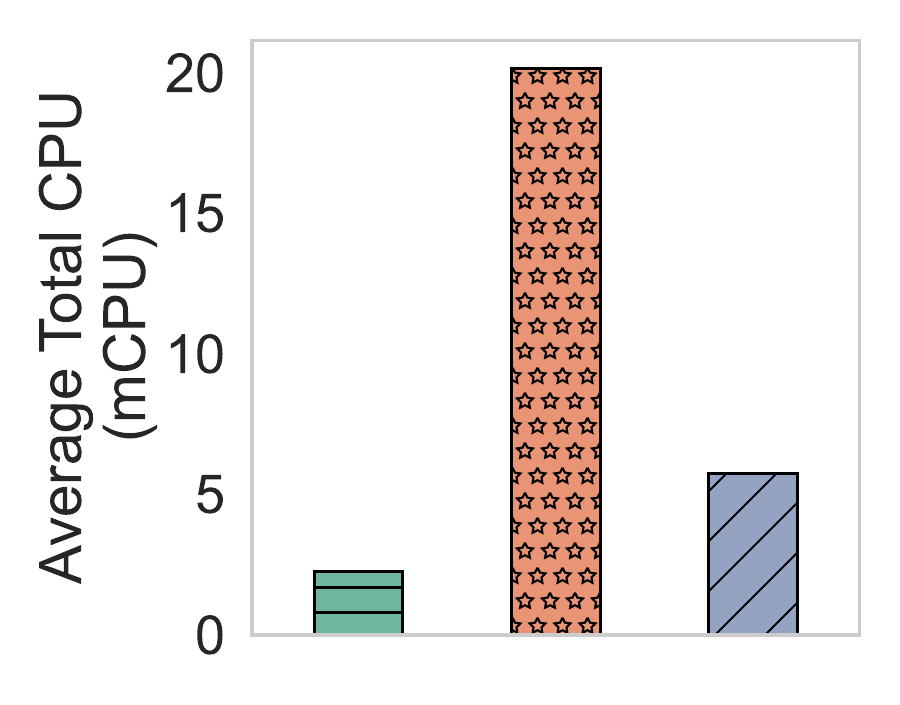}
    \label{fig:subfig3_dask_cpu}
  \end{subfigure}%
  \begin{subfigure}[t]{0.25\textwidth}
    \centering
    \includegraphics[width=\linewidth]{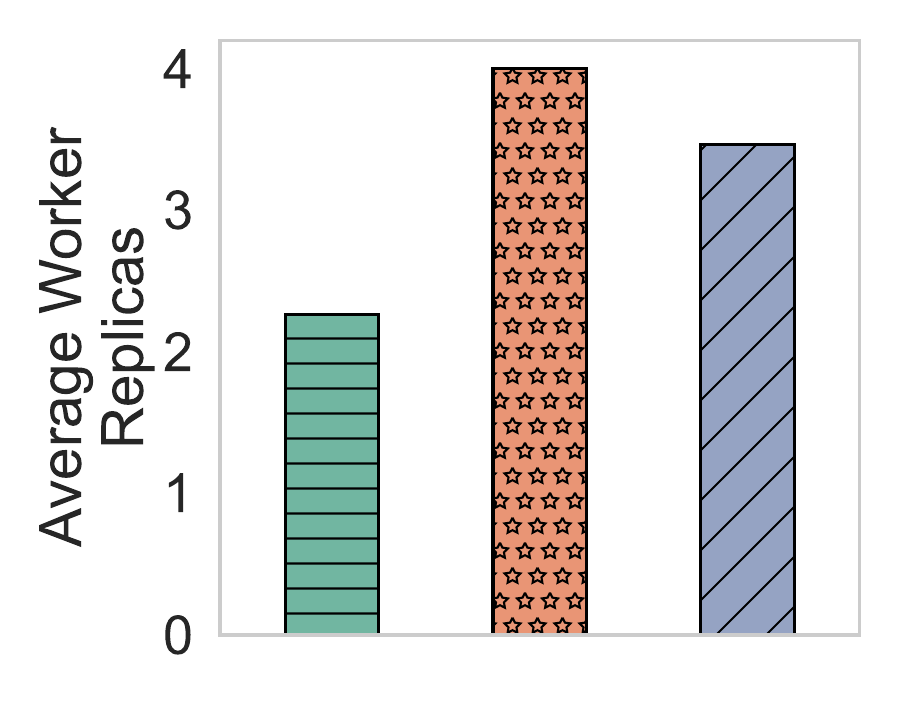}
    \label{fig:subfig4_dask_replicas}
  \end{subfigure}
  \caption{Dynamic validation of Dask over 15 different benchmarks for different User Intents. Log-scale Completion times (top left) and Benchmark Processing Costs (top right) of benchmarks, allocations of Memory (bottom left), CPU as milliCPU, or $1/1000$ of a CPU (bottom mid), and replicas (bottom right).}
  \label{fig:dask_dynamic_validation}
\end{figure*}

\begin{figure*}[ht]
  \centering
      \includegraphics[width=0.8\linewidth]{figures/legend_aws_lambda_time_comparison_dask.pdf}
    \hfill
  \begin{subfigure}[t]{0.495\textwidth}
    \centering
    \includegraphics[width=\linewidth]{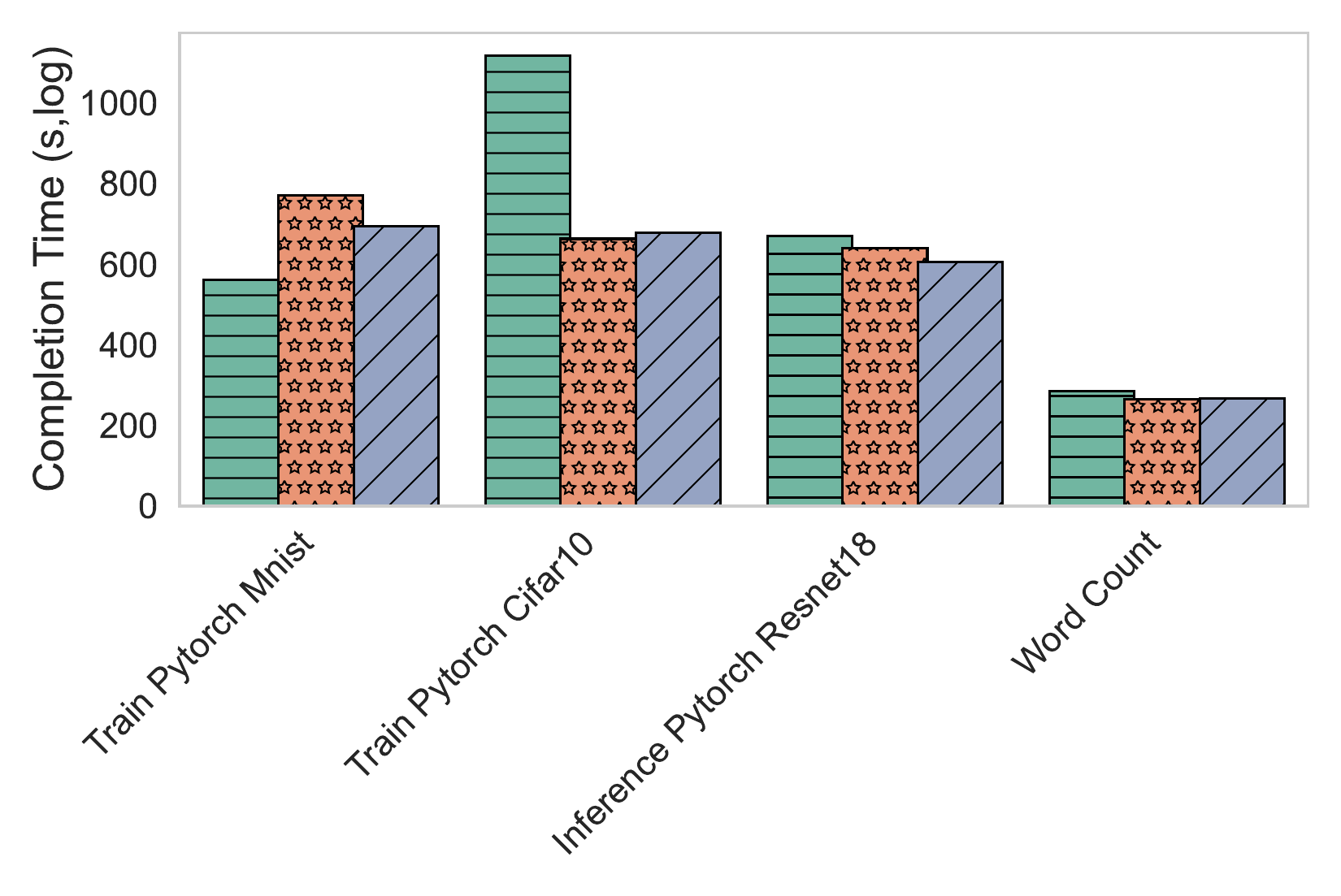}
    \label{fig:subfig1_ray_completion_time}
  \end{subfigure}%
  \hfill
      \begin{subfigure}[t]{0.495\textwidth}
    \centering
    \includegraphics[width=\linewidth]{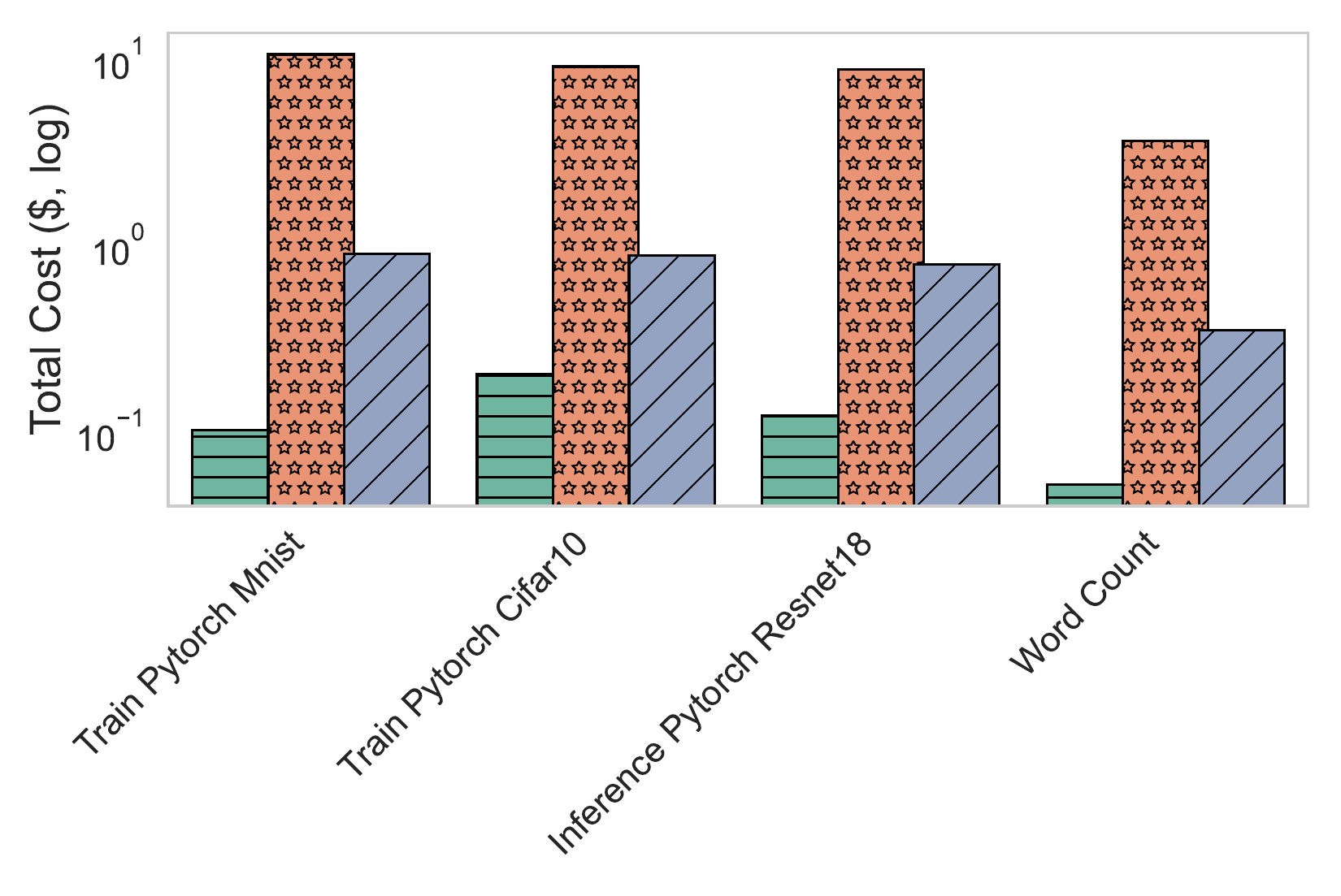}
    \label{fig:subfig1_ray_costs}
  \end{subfigure}%
  \hfill
  \begin{subfigure}[t]{0.25\textwidth}
    \centering
\includegraphics[width=\linewidth]{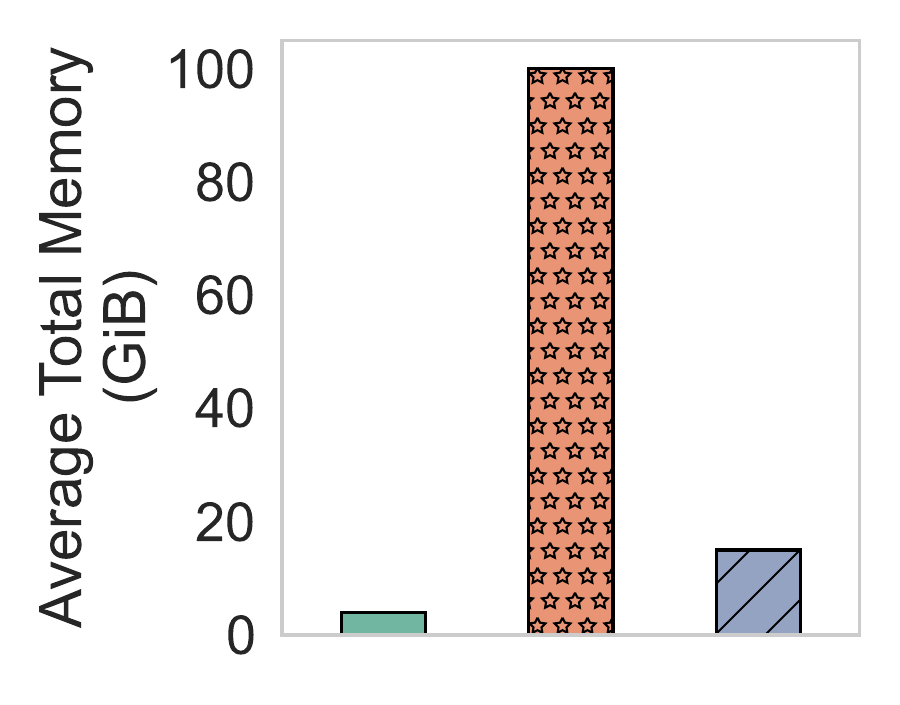}
\label{fig:subfig2_ray_memory}
  \end{subfigure}%
  \begin{subfigure}[t]{0.25\textwidth}
    \centering
\includegraphics[width=\linewidth]{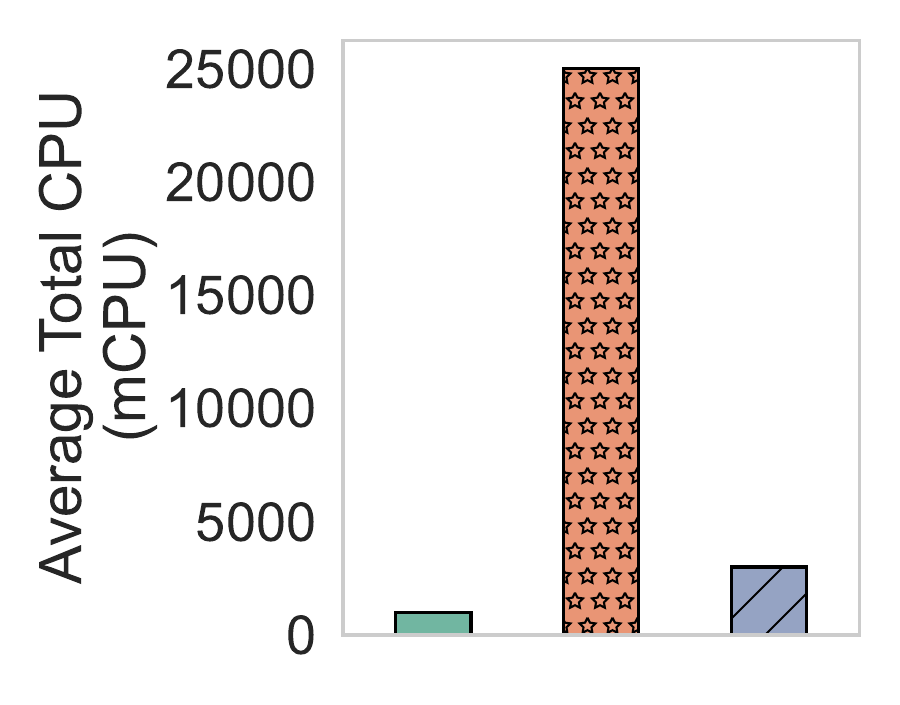}
    \label{fig:subfig3_ray_cpu}
  \end{subfigure}%
  \begin{subfigure}[t]{0.25\textwidth}
    \centering
    \includegraphics[width=\linewidth]{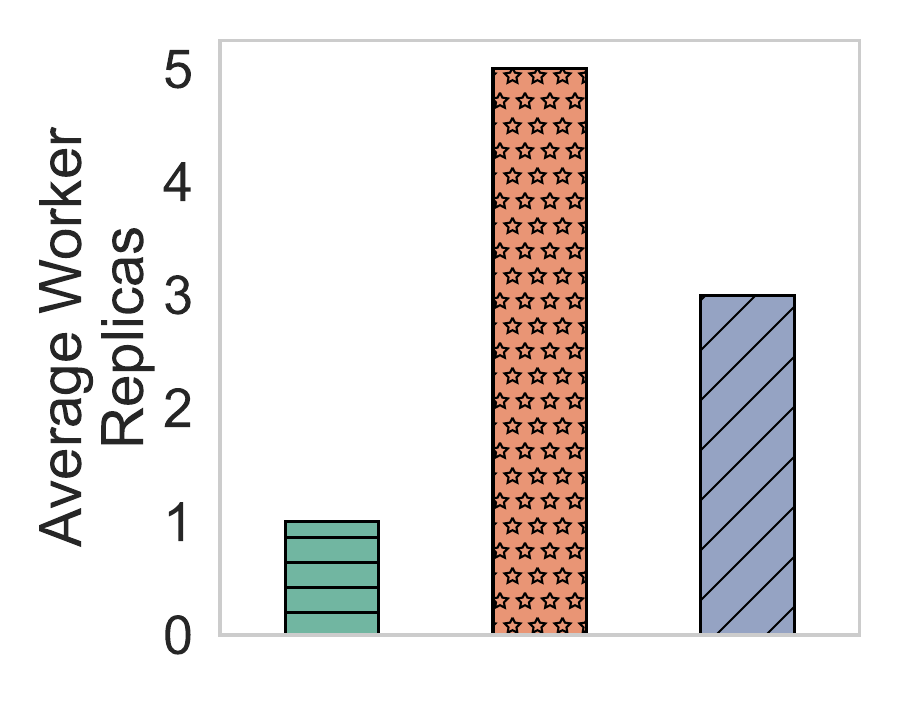}
    \label{fig:subfig4_ray_replicas}
  \end{subfigure}
  \caption{Dynamic validation of Ray over 4 different benchmarks for different User Intents. Log-scale Completion times (top left) and Benchmark Processing Costs (top right) of benchmarks, allocations of Memory (bottom left), CPU as milliCPU, or $1/1000$ of a CPU (bottom mid), and replicas (bottom right).}
  \label{fig:ray_dynamic_validation}
\end{figure*}

To evaluate the effectiveness of {\proj} in dynamic settings, we conducted extensive experiments on Dask and Ray across multiple benchmarks. We show the reults of Redis in Table~\ref{tab:main_table} in section~\S\ref{subsubsec: Dynamic Validation}. The results provide insights into the system's adaptability, resource allocation, and cost efficiency under varying workloads.

\paragraph{Dask Dynamic Validation.}
Figure~\ref{fig:dask_dynamic_validation} presents the performance of Dask across 15 different benchmarks. The top-left subfigure shows the log-scale completion times, highlighting the variations in execution efficiency. The top-right subfigure illustrates the benchmark processing costs, reflecting the economic impact of different configurations. The bottom row presents resource allocation trends, including memory requests (bottom left), CPU allocation (bottom middle), and worker replicas (bottom right). The results demonstrate that \proj effectively aligns the configuration according to user intent. Some Dask benchmarks such as ``Tasks Nearest Neighbor'' and ``Tasks Client In Loop'' are more efficient with ``High Scalability'' compared to the ``High Efficiency with balanced Scalability \& Cost'' because the latter not only prioritizes efficiency but also balances scalability and costs. This shows that \proj can successfully handle complex user intents that involve various resource and performance tradeoffs.

\paragraph{Ray Dynamic Validation.}
Figure~\ref{fig:ray_dynamic_validation} showcases the dynamic validation of Ray over four different benchmarks. The top-left subfigure compares log-scale completion times, while the top-right subfigure highlights the benchmark processing costs. The bottom row further breaks down the memory (bottom left), CPU (bottom middle), and replica (bottom right) allocations. The findings indicate that \proj successfully enhances resource utilization and cost-effectiveness, making Ray more adaptive to workload variations.

\end{document}